# evo* 2025

**The Leading European Event on**

**Bio-Inspired AI**

*Trieste (Italy). 23-25 April 2025*

# – LATE-BREAKING ABSTRACTS –

**Editors:**

**A.M. Mora**
**A.I. Esparcia-Alcázar**
**Maria Sofia Cruz**

# Preface

These proceedings include the Late-Breaking Abstracts accepted for the Evo* 2025 Conference, hosted in Trieste (Italy), from April 23th to 25th.

These extended abstracts were presented through short talks at the conference, providing an overview of ongoing research and initial results on the application of diverse Evolutionary Computation strategies and other Nature-Inspired methodologies to practical problem domains.

Collectively, these contributions point to encouraging directions for future work, underscoring the potential of nature-inspired approaches—especially Evolutionary Algorithms—for advancing research and enabling new applications.

*Antonio M. Mora*
*Anna I. Esparcia-Alcázar*
*Maria Sofia Cruz*

# Table of contents

# Effect of Fitness Values Weighting on Clustering of Single-Objective Fitness Landscapes

Vojtěch Uher[1[0000−0002−7475−3625]] and Pavel Krömer[1[0000−0001−8428−3332]]

Department of Computer Science, VSB – Technical University of Ostrava,
Ostrava-Poruba, Czech Republic, {vojtech.uher, pavel.kromer}@vsb.cz

**Abstract.** Exploratory Landscape Analysis is a powerful technique for characterizing the fitness landscapes of optimization problems. This study evaluates the use of fitness h istograms for c haracterizing b lack-box optimization problems, drawing on benchmark single-objective problems. Fitness histograms, while effective in capturing the distribution of fitness values, do not account for the relative importance of individual bins. To address this, we explore several weighting methods, particularly based on the TF-IDF statistic, to improve the histograms' discriminative power. The impact of these methods is assessed using clustering analysis, with promising results showing improved silhouette scores.

**Keywords:** Exploratory Landscape Analysis · fitness landscape · fitness histograms · TF-IDF.

## 1 Introduction

Exploratory Landscape Analysis (ELA) [4] is a versatile method used to characterize the fitness landscapes (FLs) of problems comprising a series of procedures designed to map the hypersurfaces created by the fitness values and other key characteristics of problem solutions. Studies have shown that the values of landscape features are impacted by the selection of the sampling strategy [5].

Besides the robust ELA-based feature sets [2], a novel straightforward feature based on the fitness values distribution has been introduced [6]. Each test problem / FL is represented by a normalized histogram of fitness values obtained for a sample set evaluated by its objective function. It showed good results for BBOB single-objective problems characterization and classification [6].

The normalized fitness histogram effectively captures the distribution of sample fitness values; however, it does not account for the relative importance of individual bins. The main aim of this short paper is to investigate several weighting methods to improve the distinctive properties of the fitness histograms, especially based on the TF-IDF statistic [3]. The impact of the weighting methods is evaluated using the standard clustering analysis. The experiments show promising results improving the silhouette score.

## 2   Methodology

This section explains the proposed pipeline and briefly summarizes the related methods. The pipeline can be outlined as follows: 1) Test problems/functions are chosen, 2) Sets of random samples/solutions are generated by selected strategy, 3) Samples are evaluated by test functions, 4) Feature vector is computed for each sample set in the form of a normalized histogram of fitness values, 5) The histograms are weighted using the proposed methods based on relative frequencies, and 6) used for further problem characterization, and clustering.

### 2.1   Benchmark Framework

This section briefly summarizes the test suite and methods used. The well-known 24 `BBOB` single-objective benchmark problems contained in the COmparing Continuous Optimizers (COCO) platform [1] have been selected. This study utilizes only the first instance of each problem in dimensions $d \in \{5, 10, 20\}$, and supported domain $[-5, 5]^d$. The problem landscapes are sampled using the *Sobol sequence-based sampling (Sobol)* having good space-filling properties [5]. The standard Euclidean distance is used for cluster analysis. The compactness of histograms of different `BBOB` problems is assessed with the silhouette score [6].

### 2.2   Normalized Fitness Histogram and Bins Weighting

Set of points $\boldsymbol{P} = \{\boldsymbol{p}_1, \ldots, \boldsymbol{p}_n\}$ of size $n = 2^m$ for $m = 14$ is generated in $d$-dimensional space $\mathcal{J}^d = [0, 1)^d$, $\boldsymbol{p}_i \in \mathcal{J}^d$ using the Sobol samling. Given a fitness function, $f : \mathbb{R}^d \to \mathbb{R}$, and a set $\boldsymbol{P}$, the set of fitness values is computed as $V = \{v; \forall \boldsymbol{p}_i \in \boldsymbol{P} : v = f(\boldsymbol{p}_i)\}$. The set, $V$, is utilized to compute a histogram of $h$ bins $\boldsymbol{c} = \{c_1, \ldots, c_h\}$ within the range of values $[\min(V), \max(V)]$, subject to $\sum_{j=1}^{h} c_j = n$, where $c_j$ is the number of fitness values falling to the $j$-th bin. The normalized histogram represents a discrete probability distribution of fitness values $\boldsymbol{n} = \{c_1/n, \ldots, c_h/n\}$ for $\sum_{j=1}^{h} n_j = 1$.

The normalized fitness histogram accurately represents the distribution of the sample fitness values, but it does not reflect the significance of bins. Multiple fitness landscapes can have a partially similar distribution of fitness values that on average can be numerous, and therefore, has a strong impact on the distance measures. In text mining, the TF-IDF statistic [3] is commonly used for weight calculation to determine the importance of terms within a set of documents consisting of Term Frequency (TF - term within a document) and Inverse Document Frequency (IDF - term across the set of documents).

In our case, a histogram represents a distribution of decimal values, and most of the bins contain some samples. The IDF based on the raw occurrence in the histogram cannot be used. This paper proposes a solution based on the relative frequency in the bins. Given a problem $g \in G$, where $G$ is a set of all problems, and the corresponding set of normalized histograms $N = \{\boldsymbol{n}_0, \ldots, \boldsymbol{n}_{|G|}\}$, the TF can be simply reformulated as

$$tf_H(j, g) = n_{g,j} = c_{g,j}/n, \tag{1}$$

which represents the relative frequency of the $j$-th bin in the normalized histogram representing the problem $g$. The IDF is then formalized as

$$idf_H(j, G) = \log \frac{|G|}{\sum_{g' \in G} n_{g',j}} = \log \frac{|G| \cdot n}{\sum_{g' \in G} c_{g',j}}, \tag{2}$$

which determines the weight of the $j$-th bin of all histograms and the TF-IDF is

$$tfidf_H(j, g, G) = tf_H(j, g) \cdot idf_H(j, G). \tag{3}$$

The probabilistic inverse document frequency (pIDF) [3] is added here for comparison, and it is defined as

$$pidf_H(j, G) = \log \max \left\{ 0, \frac{|G| - \sum_{g' \in G} n_{g',j}}{\sum_{g' \in G} n_{g',j}} \right\}, \tag{4}$$

and it can simply replace the $idf_H$ in equation 3. The final weighted histogram has to be normalized again as described in the beginning of this section.

## 3 Contribution

The basic question of this paper was if the TF-IDF weighting applied to normalized fitness histograms affects the representativeness of underlying landscapes which is evaluated in terms of clustering. The TF-IDF and the TF-pIDF weighting schemes were tested for the named configurations and the preliminary results can be seen in charts in figure 1 for a number of histogram bins $h \in \{8, 15\}$. Our experiments, conducted on 24 BBOB test functions using Sobol sampling showed that the TF-pIDF weighting outperformed TF-IDF, and the original normalized histograms without any bins weighting. The differences are more significant for lover dimensions while the results are even for $d = 20$. This simple scenario shows that the weighting can improve the representativeness of fitness histograms.

**Acknowledgments.** This work was supported by the Czech Science Foundation in the project "Constrained Multiobjective Optimization Based on Problem Landscape Analysis", grant no. GF22-34873K and the Student Grant System, VSB – Technical University of Ostrava, grant no. SP2025/016.

**Disclosure of Interests.** The authors have no competing interests to declare that are relevant to the content of this article.

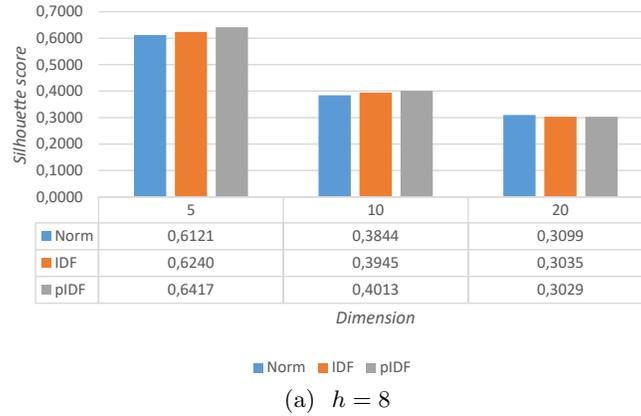

(a)  $h = 8$

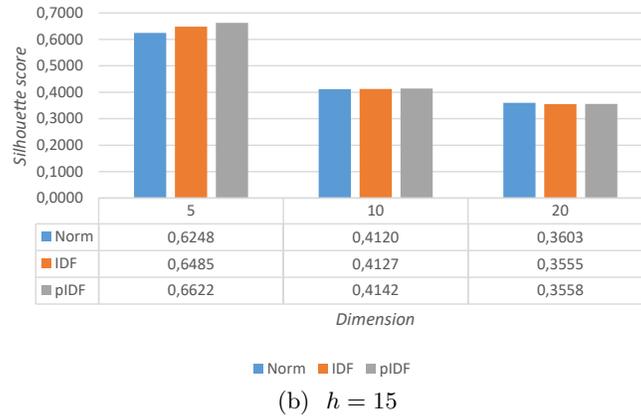

(b)  $h = 15$

Fig. 1: Silhouette score (higher is better) comparison for Sobol sampling, Euclidean distance, $h \in \{8, 15\}$, $m = 14$, and $d \in \{5, 10, 20\}$.

# The Traveling Tournament Problem: Constraint Violations for Different MaxStreak Values

Lukas Hassel[1][0009−0003−1671−4382] Kristian Verduin[2][0009−0005−8754−7635]
Bas Loyen[2][0009−0002−7935−4065] Sarah L. Thomson[3][0000−0001−6971−7817]
Thomas Weise[4][0000−0002−9687−8509] Daan van den Berg[2][0000−0001−5060−3342]

[1]KEXXU Robotics Amsterdam, The Netherlands
[2]VU University Amsterdam, The Netherlands
[3]Edinburgh Napier University, United Kingdom
[4]Institute of Applied Optimization, Hefei University, China
`daan@yamasan.nl`

**Abstract.** We systematically investigate the validity of randomly generated traveling tournament problem solutions under different values for one of its key constaints: the maxStreak, which controls consecutive game order. It turns out that the expected number of maxStreak violations closely scales with the maxStreak value, and, in the extreme case, can be even more prohibiting than any other constraint of the problem.

## 1 The Traveling Tournament Problem

The Traveling Tournament Problem (TTP) is an NP-hard problem [8] that entails scheduling tournament rounds for an even number of teams ($n_{teams} \geq 4$). The objective is to minimize the teams' travel distance, but the tournament must abide by three hard constraints [2] (also see Fig. 1):

1. Every team plays every other team *twice* in the tournament, once at home and once away, which is known as the **doubleRoundRobin** (DRR) constraint.
2. The maximum number of consecutive games a team can play at home or away, is the **maxStreak** constraint. Usually $maxStreak = 3$, meaning any team can at most play three consecutive rounds at home, or away, anywhere in the tournament.
3. When team $t_1$ plays team $t_2$ at home in one round, the inverse match ($t_2$ playing $t_1$ at home) cannot take place in the consecutive round, which is known as the **noRepeat** constraint.

As it turns out, these three innocuously looking constraints are so prohibitive, that the objective of minimizing travel distance is seldom attained. Instead, just finding a *valid* tournament at all is so hard, that the usual approach of (meta)heuristic optimization for NP-hard problems is *also* unattainable for this problem [10, 11]. Uniformly randomly sampling a valid solution appears not to be possible in subexponential time and there are no known efficient mutation or crossover operators that connect the entire search space of valid individuals [9–11]. For these reasons, we believe that the traveling tournament problem is fundamentally harder than the traveling salesman problem, even though both are classified as NP-hard. In this short study, we will investigate the number of expected violations for varying values for the maxStreak constraint.

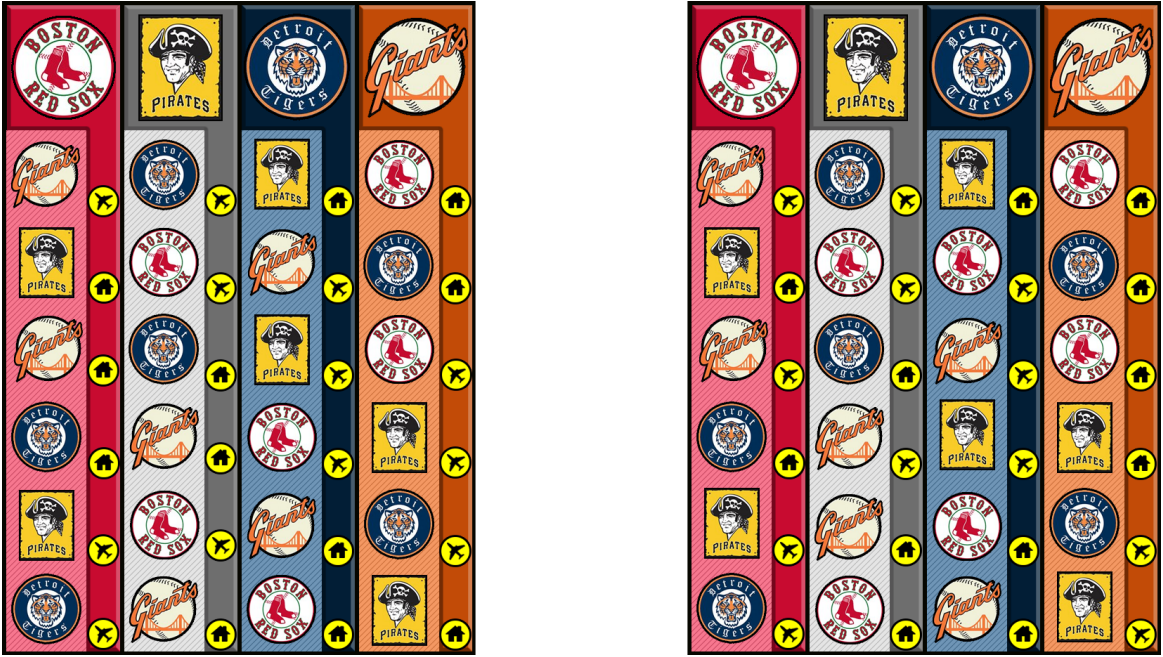

**Fig. 1. Left:** A valid 4-team TTP tournament. **Right:** An invalid TTP tournament, having a noRepeat violation in column 2, a maxStreak=3 violation in column 4, and doubleRoundRobin violations in rounds (rows) 2, 3 and 4.

## 2   Randomly Creating Tournaments

A complete tournament for the TTP consists of $2 \cdot (n_{teams} - 1)$ rounds, and in our implementation, each round is created by randomly shuffling a list of integers $[0, ..., n_{teams} - 1]$, followed by matching up the first and second integers, third and forth, and so on. As such, a round will consist of $\frac{n_{teams}}{2}$ tuples $(t_a, t_b)$, which are matches in which $t_a$ plays at home and $t_b$ plays away. Although the entire tournament is uniformly randomly filled up in $O(n)$ time, the method (necessarily?) disregards the three TTP constraints (doubleRoundRobin, maxStreak, noRepeat).

After filling up the tournament in this row-first manner, the constraint violations are counted. Every tournament must contain the games $(t_a, t_b)$ for all values $a \neq b \in n_{teams}$. For each game missing, a violation is counted, as well as for each game appearing more than once in a row. First note: in a way, this method counts each DRR-violation *twice* compared to more classical matrix-based approaches [9]. Second note: although DRR-violations are rife, they do not occur in a single round, which is guaranteed by the row-first implementation. Each round assessed in isolation is thereby valid.

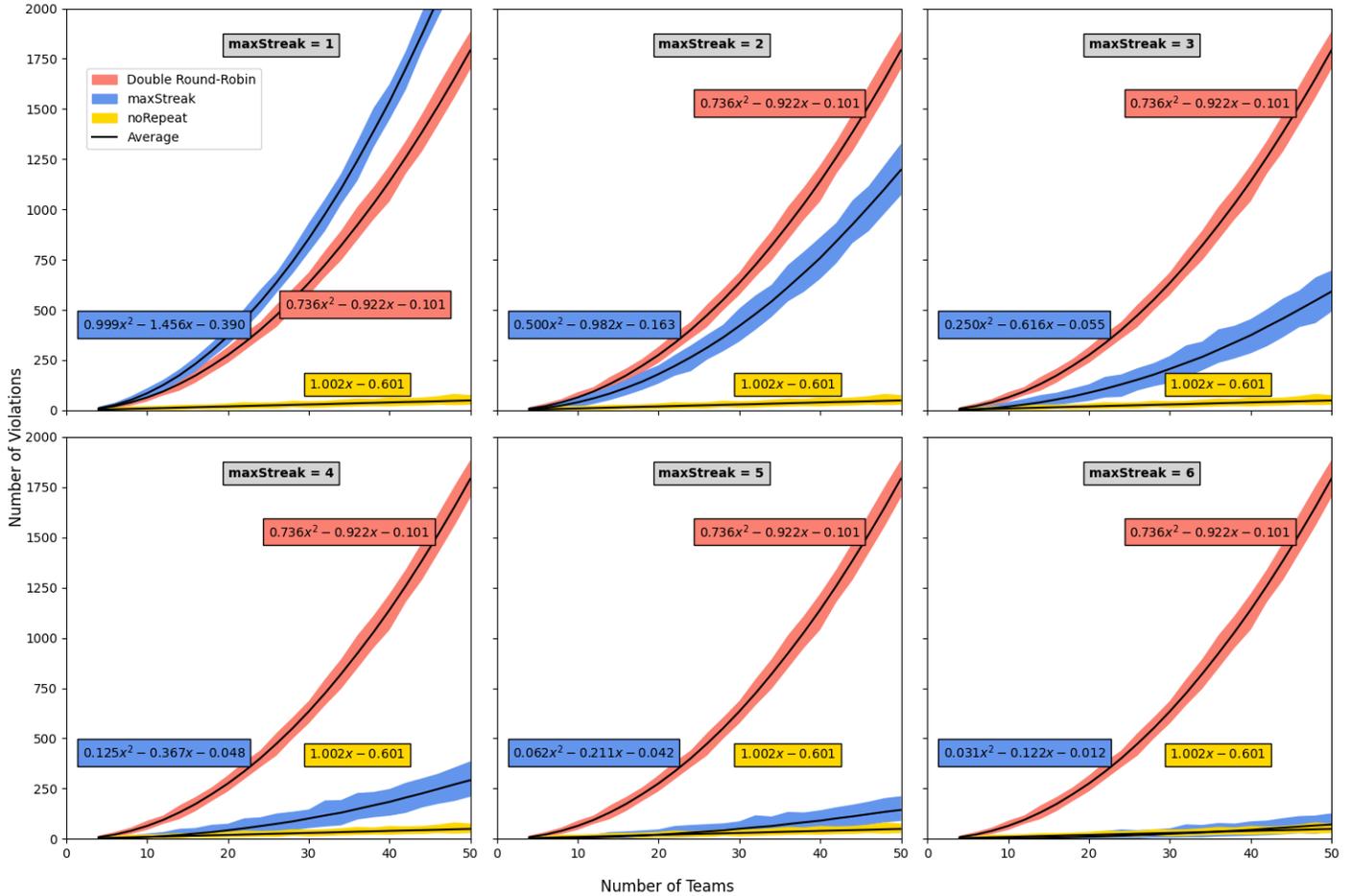

**Fig. 2.** The increase in violations for different maxStreak values.

If team $t_1$ plays more than $maxStreak$ games at home or away consecutively, each game after the preset value counts as an additional maxStreak violation on the tournament. Finally, if team $t_1$ plays team $t_2$ in round $r$ but also in round $r + 1$, this accounts for an additional noRepeat violation. Note that by using tuples for creating tournament rounds, noRepeat violation numbers are exactly half of Verduin et al.'s earlier results, which are matrix-based. For other methods, such as random permutation or pure random allocation for teams to slots, this is not necessarily the case.

In this work, one thousand random tournaments were created in the row-first manner for each $n_{teams} \in \{4, 6, 8...46, 48, 50\}$, summing up to 24,000 random tournaments, which were violation-counted for $1 \leq maxStreak \leq 6$. All work was done one an M2 MacBook Pro, and the Python source code is publicly available [4].

## 3    Results and Implications

For each $n_{teams}$, the maximum, minimum and average values for each type of constraint violation were recorded, after which we fit a quadratic function to the average (Fig. 2). For the 'default' value $maxStreak = 3$ (see e.g. [2]), our results for expected maxStreak-violations are nearly identical to earlier results [9]. But for $maxStreak = 2$, the increase in violations is *twice* as steep in $n_{teams}$. When $maxStreak = 1$, the increase is *again* twice as steep, increasing even faster than the doubleRoundRobin constraint violations. For larger values ($maxStreak = 4, 5, 6$), the increase becomes twice as *un*steep per unit increase, which leaves the summarizing characterization of the empirically expected maxStreak violations as

$$A \cdot (n_{teams})^2 + B \cdot n_{teams} + C \tag{1}$$

where A $= \frac{1}{2^{maxStreak-1}}$ and B and C are small negative constants that might to converge to zero as the value of maxStreak increases. The family of polynomials in Eq.1 demonstrates how the maxStreak parameter might be one reason that random valid solution sampling for the TTP appears impossible, which might, in turn explain the hardships of authors trying to solve the TTP with metaheuristic algorithms [1, 3, 7, 5, 12]. An immediate followup question is whether a TTP without either a maxStreak or a doubleRoundRobin constraint is *still* too hard for metaheuristic algorithms. To be honest: we think so.

## 4    Acknowledgement and disclaimer

Figure 1 serves an illustrative purpose; it cannot be made by our algorithm. In fact, we suspect that any sensible row-first algorithm cannot create invalid tournaments with 1,2 or 3 noRepeat violations for $n_{teams} = 4$. Logos in this figure were remade by melling2293@Flickr and are distributed under creative commons license. This is a twin submission, together with [6].

# The Traveling Tournament Problem: Valid Solutions are Very Different

Bas Loyen[1][0009−0002−7935−4065] Lukas Hassel[2][0009−0003−1671−4382]
Sarah L. Thomson[3][0000−0001−6971−7817] Thomas Weise[4][0000−0002−9687−8509]
Daan van den Berg[1][0000−0001−5060−3342]

[1]VU University Amsterdam
[2]KEXXU Robotics Amsterdam
[3]Edinburgh Napier University, United Kingdom
[4]Institute of Applied Optimization, Hefei University, China
daan@yamasan.nl

**Abstract.** We generated and rendered all 160 valid solutions for the 4-team traveling tournament problem, and they look nothing alike. The average differences between two solutions is almost maximal, even when when a key constraint, the home/away designation, is completely ignored. If these findings hold for larger numbers of teams, it could prohibit the use of metaheuristic algorithms for this problems altogether.

## 1 The Traveling Tournament Problem

The Traveling Tournament Problem (TTP) is an NP-hard problem in which a tournament for an even number of teams ($n_{teams}$) should be scheduled [1]. The goal is to minimize the teams' travel time between venues, and the tournament should abide by three constraints (See Fig. 1):

1. Every team plays every other team *twice* in the tournament (once at home, once away), which is known as the **doubleRoundRobin** constraint.
2. When team A plays team B in one round, the inverse match (B playing A) cannot take place in the consecutive round, which is known as the **noRepeat** constraint.
3. The maximum number of consecutive games a team can play at home or away, is the **maxStreak** constraint. Usually, $maxStreak = 3$ meaning any team can at most play three consecutive rounds at home, or away, anywhere in the tournament  [7].

These three constraints are so prohibitive, that a number of vexing complications arise. Just finding a *valid* solution is already so hard, that the usual approach of (meta)heuristic optimization for NP-hard problems appears unattainable. Uniform random sampling of valid solutions appears not to be possible in subexponential time and, possibly related, there are no known efficient mutation or crossover operators that connect the entire search space of valid individuals [10, 8, 9, 11]. For these reasons, we believe that the traveling tournament problem is substantially harder than the traveling salesman problem (TSP), even though both are classified as NP-hard. In this short study, we will suggest that the unavailability of these operators might be due to the dissimilarity between valid schedules.

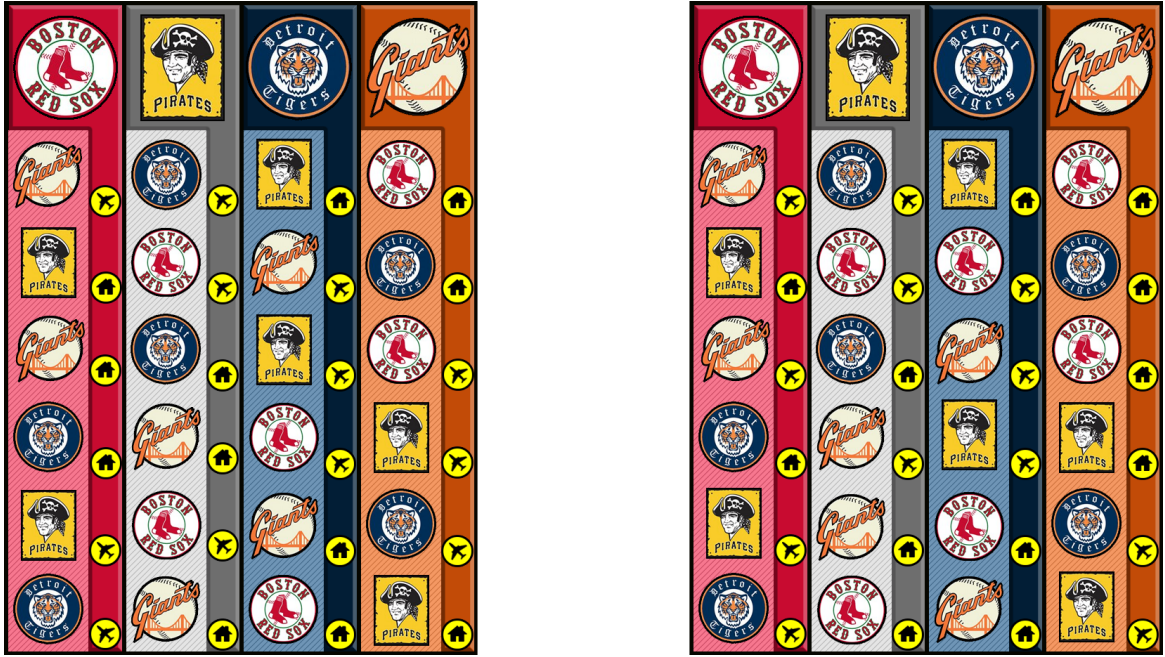

**Fig. 1. Left:** A valid 4-team TTP tournament. **Right:** An invalid TTP tournament, having a noRepeat violation in column 2, a maxStreak=3 violation in column 4, and doubleRoundRobin violations in rounds (rows) 2, 3 and 4.

## 2    Finding all 160 valid solutions

A complete tournament for the TTP with $n_{teams} = 4$ consists of $2 \cdot (n_{teams} - 1) = 6$ rounds. Assuming every round holds every team exactly once (which seems sensible), the number of testable tournaments would amount to $(n_{teams}!)^{2(n_{teams}-1)}$ which for $n_{teams} = 4$ still accounts for over 191 million tournaments – an impractically large number. Since we require *all* valid tournaments, the reasonable way forward appears to be depth-first search with runtime pruning, which can indeed make quite a difference for NP-complete problems [5, 6, 4].

The first round of the tournament always holds matchups $(t_0, t_1)$ and $(t_2, t_3)$, the first of each tuple playing at home. Fixing the first round in this 'half-normalized' way breaks symmetry by a factor $\frac{(n_{teams})!}{(1/2 n_{teams})!} = 12$. For the following rounds, the depth-first algorithm recursively tries all combinations of matchups for each round, but after the placement of any single matchup however, checks whether any of the 3 constraints in the partial schedule are already violated. If so, the search on that subbranch is halted and the algorithm backtracks, at most until just before the first round, which is fixed. This process produces all $n_{schedules} = 160$ half-normalized valid TTP solutions, after which we calculate their $^1/_2(n_{schedules})(n_{schedules} - 1) = 12,720$ mutual differences. Between any two of these schedules, at most $6 \times 4 - 4 = 20$ slots can be different, either by team placement or by home/away designation. As such, the difference between any two tournaments yields an integer between 0 and 20. In Figure 2, the top window shows these differences, which are very high ; an average of $\mu = 16.63$ entries out of the 20 variable slots in the tournament are different. But even when we completely ignore the home/away designations (central window), the average difference is still $\mu = 14.49$. Even when we consider the bare skeleton of home/away designations *only*, without team placements (bottom window), still an average of $\mu = 9.50$ slots are different. That is still almost half the schedule for this vast constraint relaxation.

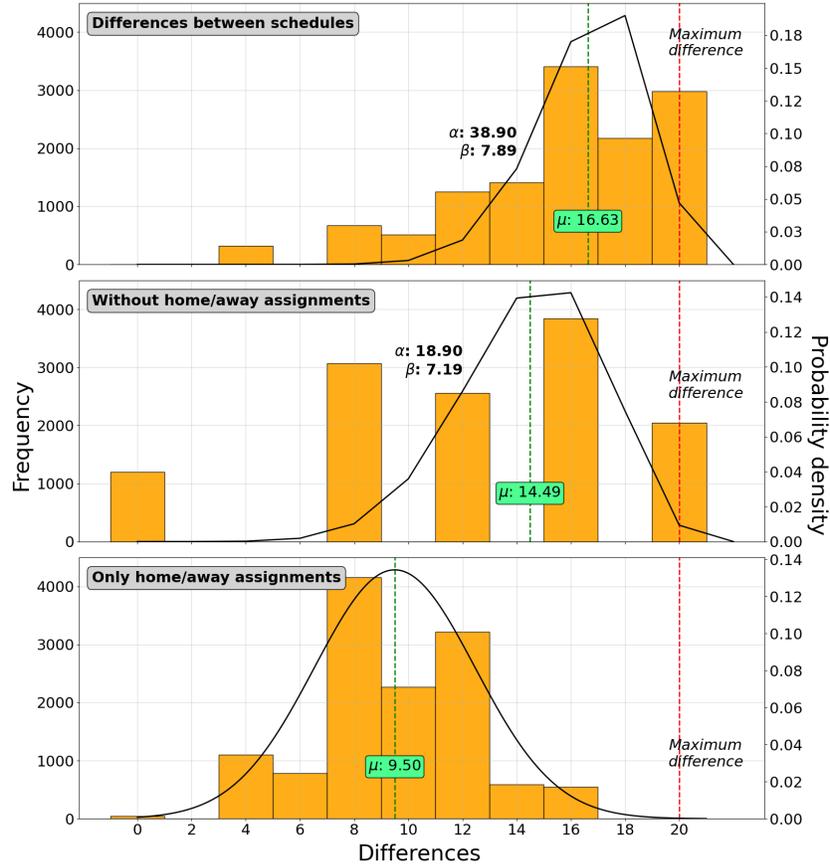

**Fig. 2. Top:** Differences between valid tournaments for $n_{teams} = 4$ are close to maximal. But even when home/away designations are ignored (**middle**) or even team placement is ignored (**bottom**), differences are still high.

## 3   Implications

The consequences of the near-maximal difference between two valid solutions of the traveling tournament problem for $n_{teams} = 4$ is potentially catastrophic for its solvability. At the very least, it means that potential mutation and crossover operators are nontrivial and must make enormous changes from one valid schedule to the next. It is even conceivable that no realistic, short-formulation mutation operators exist, or even when they do, that they do not connect all valid individuals.

This is a sharp departure from other NP-hard problems like the traveling salesman problem, that has a host of common-sense mutation operators, most of which have short formulations and effortlessly connect all valid individuals. Therefore, the results could be seen as evidence that the TTP is much harder than the TSP – even though both are listed in the same hardness class. How should we resolve this issue?

## 4   Acknowledgement and disclaimer

The right hand side of Figure 1 serves an illustrative purpose; we suspect that any sensible row-first algorithm cannot create invalid tournaments with 1,2 or 3 noRepeat violations for $n_{teams} = 4$. Python source code, and auxiliaries to this publication are publicly available [3]. Logos in this figure were (re)made by melling2293@Flickr and are distributed under creative commons license. This is a twin submission, together with [2].

# AI as Co-Creator: Reimagining Musical Authorship and Interaction in Human-AI Music Composition

Chen Wang

[1] University of the Arts London, London SE1 6SB, UK
c.wang1220212@arts.ac.uk

**Abstract.** This paper explores how human-AI collaboration reshapes creativity, authorship, and composition through *Cyber Maze*, a practice-based project where AI serves as a generative partner. Rather than functioning autonomously, AI introduces unpredictable sonic variations that challenge fixed creative paths and prompt novel directions in music-making. In this iterative process, the composer's role evolves from sole author to curator and facilitator, navigating a dynamic interplay between algorithmic suggestions and artistic decisions. Vocal recordings and musical structures are continuously transformed through human-machine exchanges, resulting in a fluid compositional model. This study contributes to ongoing discussions in AI-assisted music by highlighting collaborative frameworks that foreground adaptability, nonlinearity, and creative negotiation between artists and computational systems.

**Keywords:** AI Music Composition, Human-AI Collaboration, Generative Music, Computational Creativity, Creative Agency in AI, Openness in Music Creation.

## 1    Introduction

The integration of artificial intelligence (AI) into music production has transformed how artists engage with sound, composition, and authorship [1] [2] [3] [4]. From algorithmic lyric generation to AI-assisted arrangement tools, AI is no longer simply a means of automation but a participant in shaping creative outcomes [5] [6] [7]. This shift calls into question traditional understandings of artistic agency, raising concerns and opportunities around how musical ideas are conceived, developed, and credited.

This paper investigates these questions through Cyber Maze, a creative project in which AI acts as a generative collaborator in music composition. By iteratively engaging with AI-generated material—selecting, modifying, and reinterpreting its outputs—the composer engages in a nonlinear workflow, where intention and surprise continuously inform each other. The project seeks to understand how AI reshapes creative roles and to what extent it influences the direction, structure, and expression of the final musical work.

## 2    Background and Related Works

### 2.1    The Expanding Role of AI in Music

AI's role in creative music practices has expanded significantly in recent years. From melody generation to vocal synthesis and harmonic structuring, AI tools such as Suno [8] and FLOW Machines [9] allow composers to co-create in ways that challenge fixed roles in traditional workflows. These systems generate musical ideas based on training data, but their unpredictability and generative nature open up space for experimentation, particularly when composers engage critically with the machine's outputs [10].

In this evolving relationship, AI is less a tool of precision and more a partner in ambiguity—offering suggestions that require interpretation, curation, and transformation. As such, human-AI co-creation marks a shift from automation to collaboration, where the aesthetic and structural outcomes of music emerge through interplay rather than isolated authorship.

### 2.2    AI as a Creative Partner in Contemporary Practice

Recent music projects highlight how artists increasingly work with AI as an expressive collaborator. Holly Herndon's PROTO [11] incorporates AI-generated vocal textures trained on her own voice, while YACHT's Chain Tripping [12] involved training models on the band's previous recordings to produce recombined lyrical and melodic material. In both cases, artists acted as mediators, shaping the AI's output through selection and refinement.

Similarly, Arca's Riquiquí [13] involved generating dozens of AI-produced variations from a single composition, later curated into a final work. These examples reflect a broader trend where AI contributes to the creative process not by replacing the human composer, but by unsettling habitual patterns and offering new sonic directions. In doing so, these collaborations blur distinctions between creation and curation, control and improvisation.

### 2.3    Debates on Creativity and Authorship

The involvement of AI in music raises unresolved questions about creative agency. While some argue that AI merely recombines data and lacks artistic intention, others emphasize that recombination and variation have long been central to human creativity [14] [15]. Projects like the AI-assisted completion of Beethoven's Tenth Symphony [16] exemplify the tension—where algorithmic generation and human judgment converge in the final product.

Rather than focusing on whether AI is "truly creative," this study approaches creativity as a process of interaction. Cyber Maze frames authorship as distributed across human and machine contributions, where aesthetic meaning arises through intervention, layering, and ongoing reinterpretation.

## 3     Human-AI Collaboration in Creative Practice

The use of AI in music composition introduces a unique kind of generative unpredictability, distinct from improvisation or performer-led variation [17] [18]. AI systems, driven by algorithmic processes rather than human spontaneity, generate sonic materials that often defy initial expectations [19] [20]. These responses can exceed or undermine human intention, compelling the composer to constantly negotiate between structured vision and emergent sonic outcomes [21]. This feedback loop transforms composition into a dialogue—one where control is distributed, and creative decisions are continuously re-evaluated in response to machine outputs.

Such interaction redefines the role of the composer, shifting it from originator to facilitator and curator. In the development of Cyber Maze, this shift was palpable. AI-generated sounds often catalyzed new compositional directions, requiring the composer to remain flexible and receptive to disruption. Iterative reuse of AI-generated motifs, fragments, and textures became central to the process, turning composition into a recursive cycle of experimentation and reinterpretation. Moreover, the AI's ability to introduce unfamiliar stylistic material prompted continual reassessment of form and structure, fostering an exploratory mindset that valued surprise, openness, and deviation from habitual patterns.

## 4     Methods

The compositional workflow of Cyber Maze was deliberately constructed around iterative interaction with AI systems. Text-based models such as GPT [22] were employed for lyric generation, using them not for complete verses but as ideation tools to inspire poetic fragments. These fragments—evocative lines like "Neon dreams and virtual fears"—were extracted and recontextualized, serving as conceptual anchors rather than definitive texts. This allowed lyrical development to remain fluid and exploratory.

In parallel, musical material was developed using Suno AI, a generative audio platform prompted with stylistic descriptors such as "hyperpop," "granular," or "psychedelic." Rather than accepting outputs as finished compositions, the process emphasized selection, recombination, and transformation. These audio results were layered, modified, and reconstructed within a DAW, echoing collage techniques that fused multiple textures and fragments into a coherent soundscape.

A key methodological feature was the recursive interaction between human vocal improvisation and AI reprocessing. Vocal ideas recorded in response to AI instrumentals were subsequently fed back into the system with varied prompts, producing a circular exchange of influence between human and machine. This cut-up and remix approach resulted in a richly textured composition shaped through dialogic iteration and digital bricolage.

## 5      Results

Cyber Maze emerged as a non-linear, evolving composition that blurred traditional creative boundaries. The AI-generated materials, when treated not as endpoints but as generative triggers, enabled a layering of aesthetic references—bridging mainstream sonic aesthetics with experimental textures. By adopting a fragmented, feedback-oriented workflow, the final composition reflected both the constraints and affordances of generative AI systems.

Crucially, the project demonstrated that while AI-generated music often conforms to patterns present in its training data, true novelty emerged through human intervention: through selection, recompositing, and interpretative shaping. The process allowed for hybrid musical forms to take shape—ones that could not have been planned in advance but unfolded through real-time navigation of AI unpredictability. The composition, thus, became a site of emergent authorship, shaped not by a single creative vision but by layered decisions responding to shifting machine outputs.

## 6      Discussion

This project reveals how AI serves less as an autonomous creator and more as a catalyst that disrupts and expands creative practice. While AI systems like Suno or GPT lack contextual understanding, cultural sensitivity, or emotional depth, they introduce sonic ideas that prompt reevaluation and reinterpretation. The AI's inability to respond dynamically to evolving artistic intent means that co-creation must be understood as a one-sided dialogue—one in which the human composer bears the responsibility of shaping, curating, and contextualizing machine-generated content.

A core tension emerged around the question of innovation versus imitation. While AI introduced novel textures and formal disruptions, it frequently reproduced familiar patterns, especially in pop and electronic genres. These patterns, while stylistically coherent, were often limited in depth, reinforcing the necessity of human agency in creating meaningful compositional arcs. Thus, authorship became a matter of shaping potentialities rather than originating materials—challenging traditional notions of creative control and linear authorship.

Ultimately, the composer's role transformed from being a generator of fixed ideas to a responsive agent navigating emergent possibilities. AI destabilized compositional hierarchies, compelling the composer to shift from dictator to explorer. This shift holds broader implications for how we conceptualize creative agency in an age of generative tools—not as static authority but as dynamic negotiation with non-human systems.

## 7      Conclusion

The case of Cyber Maze demonstrates that human-AI co-creation in music is not about automation or substitution but about rethinking the nature of creative agency. Generative AI systems, by virtue of their unpredictability, compel artists to embrace iterative workflows, curatorial mindsets, and dialogic experimentation. Rather than yielding polished results, AI introduces ambiguity, disruption, and complexity—qualities that, when critically engaged with, can become engines of innovation.

The value of AI in music lies not in its ability to independently create but in its potential to destabilize habitual practices and provoke new aesthetic responses. The composer, in this context, becomes less a sole author and more a co-navigator—interpreting, reshaping, and responding to machine-generated stimuli. This dynamic interplay redefines authorship as an ongoing process of selection, transformation, and reinterpretation, revealing new models of artistic collaboration in the age of computation. As creative roles become more fluid, the future of music-making may lie not in mastering machines, but in learning how to meaningfully collaborate with them.

# Statistical-Mechanical Approach to Music:
# A Nature-Inspired Model for Rule-Free Composition

Hamid Assadi and Sam Cave

Brunel University of London, Uxbridge UB8 3PH, United Kingdom

**Abstract.** Music is a universal language, yet its fundamental ingredients are not objectively established. This paper introduces a statistical-mechanical framework for isolating and testing candidates for these ingredients. Music is modelled as an ensemble of time-frequency events, in analogy with materials as ensembles of atoms, characterised by the macro-properties of energy and entropy. We define energy as a measure of temporal dissonance or tension, and entropy as a measure of unexpectedness or surprise; two quantities that fluctuate over time and give rise to emotionally perceptible musical contours. The model demonstrates that music-like structures can emerge outside equilibrium, without relying on predefined rules or learned styles. Yet they conform to well-established subjective norms, such as those delineated by *Cantus Firmus* and, indeed, the melodic principles of Species Counterpoint more generally. Unlike mainstream AI-generated music, which relies on trained models and probabilistic interpolation, our system is *ab initio*, operating purely through algebraic transformations in the time-frequency domain, without predefined scales, chords, or rules. This shifts composition from the frequency-time domain to the tension-surprise domain, providing a more direct and accessible connection to emotional experience. The system contains no trainable parameters, producing entirely novel, unique, and genre-fluid compositions, rather than interpolating between existing musical data; thus, mitigating copyright risks commonly associated with AI music. Beyond offering insight into the nature-inspired mechanisms underlying musical emergence, the system functions as a 'smart' instrument, enabling real-time adaptability. This makes it particularly well-suited for applications in EEG-coupled neurofeedback, music therapy, gaming, and interactive media, where dynamic emotional expression is crucial. In contrast to common generative approaches that emphasise structure or style, compositional decisions using this system are made considering instantaneous variations of the emotionally relevant physical analogues. This may offer new tools for researchers and composers seeking expressive means beyond conventional theory or machine-learned imitation.

**Keywords:** Rule-Free Composition, Music Modelling, Emotion-to-Music.

## 1 Introduction

How might we explain the concept of music to an advanced extraterrestrial, one equipped to perceive sound and reason mathematically, yet entirely unfamiliar with human culture or musical tradition? If music were wine, what would be the alcohol, the level of acidity or sugar content? These questions can have no definite answers if music is conceived as a purely subjective human-centred experience, simply because neither mathematics nor chemistry is a matter of human judgment. Likewise, the discovery of alcohol was not primarily concerned with flavour or subjective preference, but rather with identifying and isolating a specific chemical compound responsible for a known effect. On the other hand, attempts to analyse music have predominantly focused on stylistic conventions, historical developments, cultural context, emotional impact, personal interpretation, and many other human-centred aspects based on the notion that music cannot be understood outside human perception [1-4].

Like isolating active compounds in complex substances, this work seeks to identify objective quantifiable properties of music independent of stylistic rules, traditions or any factors tied to the listener's subjective response to sound. We do not aim to reduce music to those quantified properties, nor do we aim to study the effect of those properties on humans within the scope of this paper. Instead, we propose a framework for isolating and testing properties that could arguably be considered within the key carriers of music's emotional content. As ethanol is an active component that gives an alcoholic beverage its intoxicating effect, we take features such as musical tension and its temporal variations as the fundamental drivers of music's emotional and cognitive impact.

We also propose that those features can be formulated and quantified objectively using statistical-mechanical concepts. The goal is to demonstrate how these nature-inspired properties can contribute to shaping musical experience in a way that transcends style, genre, or cultural conventions. By decoupling music generation from historical precedent and learned patterns, we seek a new way to understand, analyse, and generate music as an emergent, self-organising phenomenon. We test our proposed conjecture by providing examples of generative compositions that do not explicitly follow empirical musical rules yet can be recognised as embodiments of what we refer to as music. It should be noted that using terms such as 'energy' and 'entropy' is not intended to suggest literal physical quantities but as an analogy. Musical tension and surprise are treated as macroscopic descriptors of a system composed of discrete tonal events, much like energy and entropy that describe the macroscopic properties of an ensemble of particles in thermodynamics. The goal is not to equate music with physical matter but to borrow the formalisms that describe emergent properties from local interactions.

Our motivation is twofold: (1) to provide additional insight into the nature of music and (2) to facilitate creativity without diminishing the artist's role in the process, hence moving beyond what current traditional composition models and mainstream AI music can offer. Whether human-driven or AI-based, the existing models operate within rules or constraints, be it Western music theory, jazz improvisation patterns, or deep learning models trained on past compositions. AI music, in particular, relies heavily on data-driven models trained on vast amounts of existing compositions and is hence statistically aligned with the prior art. This is the central limitation of AI music: it operates within the space of familiar patterns and what is already known. As a result, AI-generated music today functions much like a music recommendation system rather than a genuine creative engine, despite being sufficiently sophisticated at style emulation to pass the Turing test [5, 6].

By contrast, we aim to propose a generative system that does not rely on existing musical data or stylistic imitation but one building on quantifiable macro-properties that can reasonably be considered key emotion-carrying agents. A key feature of such a system is that it should be self-training in real time. Thus, the generated music may be regarded as an elementary synthesis of expressive variables, much like a simple aqueous ethanol solution. Nevertheless, it may still serve as evidence to validate the initial hypothesis, if the outcome is regarded as music by any listener. Also, in contrast to data-driven approaches, we do not treat music as a fixed body of learned structures but rather as a dynamic process shaped by the instantaneous values of the emotionally relevant macro-properties and their rates of change. This idea resonates with classical compositional practice, particularly with the harmonic rhythm (rate of change) of functionally tonal cadential progressions as exemplified in classical period phrase structures by composers such as Mozart and Haydn. It is also supported by cognitive theories of music perception and expectation [2, 7, 8]. The generated scores can in this way be a real-time reflection of the composer's emotional prescriptions, which can also serve as original seeds for subsequent fully developed musical compositions.

The foundation of our approach lies in the assumption that melodic and harmonic tension besides unexpectedness (surprise) are two key factors governing musical experience. This approach aligns with recent findings in music cognition and neuroscience, which suggest that the brain's response to music is shaped by fluctuations in tension and surprise rather than adherence to stylistic conventions [9]. We extend previous work on information theory in music [7, 10] and music perception models [8] by introducing a self-training, entropy-energy-guided system for music generation. Entropy has already been widely used in music as a measure of unexpectedness, i.e., the degree to which a listener can or cannot predict what comes next. It has been used in information-theoretic models of music cognition, where studies show that higher entropy leads to more profound perceptual uncertainty, and in many cases heightened engagement [11]. Energy, by contrast, is a less established concept in music and has been used as a general term to describe the intensity of musical events influenced by dissonance, dynamics, rhythmic drive or harmonic instability [12, 13]. In the context of the present work, energy is taken to relate to how strongly the listener feels tension or resolution. From a psychological perspective, we consider energy and entropy as intrinsic macro-properties of sound that influence valence and arousal as two main components of emotional affect in humans [14, 15]. While the type and level of influence depend on the culture and many other subjective human-centred factors, the properties can be quantified objectively regardless of human judgment.

This paper presents a framework for quantifying these properties using the same statistical mechanical concepts applied to an ensemble of atoms to derive their macro-properties. A physics-based approach to music analysis traces back to Hermann von Helmholtz [16] long before the information theory was formalised by Shannon [17]. A pioneer of thermodynamics, Helmholtz applied principles such as energy, resonance, and dissipation to explain tone perception and the sensory basis of consonance, anticipating later distinctions between order and randomness in sound. More recently, Berezovsky [18] proposed a statistical mechanics model to explain the emergence of the 12-tone equal-tempered system as a natural outcome of entropy-energy trade-offs in musical harmony. While sharing a similar physical intuition, both approaches remain primarily analytical, not offering a generative or algorithmic framework. In contrast, this work extends thermodynamic reasoning into a generative model that treats musical tension and surprise as compositional variables, enabling the creation of original music without stylistic priors or human-derived rules. The work offers a third approach besides rule-based or data-driven systems. It derives musical structure directly from emotion-relevant, physics-inspired macro-properties. It also contrasts evolutionary algorithms that optimise fitness functions or style-mimicking models that interpolate between known compositions, thus, filling a gap in the functional classification of music generation systems [5].

## 2 Methodology

The implementation of the proposed model of music (referred to here as *Tonamic*) is divided into two parts: (1) the analysis of music based on thermodynamically inspired macro-properties and (2) the real-time generation of music via entropy-energy-guided optimisation. The methodology is entirely *ab initio* and does not rely on data-driven learning, rule-based grammar, or musical examples – the only assumption is that the frequency domain is quantised into the twelve-tone system as a 'natural' choice of temperament [18].

### 2.1 Analysis of Music

The analytical component is based on a statistical mechanics-inspired model, in which tones are considered microscopic entities. The input score is tokenised using the shortest note value as the temporal unit. From these discrete sound events, macroscopic properties, specifically entropy and energy, are computed in a moving window of a fixed length along a melodic sequence as follows.

- **Entropy (S)** is calculated as a measure of the unpredictability of tone occurrences in a sequence, considering both short-range and long-range order of the ensemble of pitched events in the moving window visualised as green in the plots. Entropy is computed as a weighted combination of three Shannon entropy measures: $S = \alpha S_{\text{pitch}} + \beta S_{\text{interval}} + \gamma S_{\text{lagged}}$, where $\alpha$, $\beta$, $\gamma$ are adjustable weights, and $S_i = -\sum_i p_i \log p_i$ in which $p_i$ represents the empirical probability of pitch, interval or lagged note pairs.
- **Energy (E)** is calculated as a dissonance potential function between all active tones in the temporal ensemble. Drawing on models from Euler's *gradus suavitatis* (Euler's measure of consonance) and more recent harmonicity indices [19], the *Tonamic* energy metric incorporates both vertical (simultaneous) and horizontal (sequential) tone relationships. In analogy with concepts of the potential and kinetic energy of the physical matter, the *Tonamic* energy accounts for the interaction of the microscopic entities both with one another and with a reference ground state (tonic). The energy E (shown in blue) is modelled as a (physics-inspired) Hamiltonian-like potential function capturing dissonance between all tone pairs in the moving window, with adjustable weights for the potential and kinetic components.

The energy also includes an adjustable contribution from pitch. Both energy and entropy are dimensionless and normalised between 0 and 1. Several other auxiliary macro-properties are calculated alongside energy and entropy, including an energy gradient term accounting for the distance between consecutive tones. The property shown in red is a normalised ratio between energy and entropy. While this quantity is conceptually comparable to 'temperature' in thermodynamics (energy per unit entropy), it is not treated as a physical temperature but rather a musical analogue representing tension under uncertainty. It should be noted that formulating temperature in thermodynamics assumes the presence of equilibrium, a condition wherein energy is freely exchanged among entities until reaching a stable state. However, there is no rational basis to consider that musical systems typically operate under such equilibrium. In an isolated system, equilibrium corresponds to the macro-state with the highest number of micro-states, i.e., the maximum entropy. In a system under constant pressure and temperature, equilibrium is designated by the minimum free energy. Since no analogue of pressure exists in this musical framework, and an ensemble of random notes does not constitute music, these concepts are not meaningfully transferable to a musical context. Moreover, musical tones do not interact or exchange energy freely as particles do in a physical system. In music, the composer or the system constraints determine note occurrence and progression, which defies the notion of thermodynamic equilibrium. In other words, unlike physical systems, musical states are externally prescribed and do not naturally evolve toward equilibrium. They fluctuate between two extreme states: (1) a musical sequence composed of stable, consonant intervals and predictable patterns corresponding to a low-energy, low-entropy state, analogous to a crystalline solid, and (2) a chaotic ensemble of dissonant and unpredictable tones corresponding to a high-energy, high-entropy state, analogous to a hot gas.

## 2.2 Music Generation via Affect-Prescribed Composition

In contrast to systems trained on musical corpora, the generation algorithm is based on prescribing target trajectories for the tonal macro-properties introduced above. In this reverse process, the aim is not to imitate music but to form emergent sequences of musical characteristics as a function of their underlying property landscape. The key steps in the generative algorithm are as follows.

- **Input Prescription:** The user defines temporal target values for entropy, energy, and other optional auxiliary properties. Depending on the composer's intended emotional landscape, these curves may be sinusoidal, stepwise, or more complex patterns.
- **Initialisation:** An initial uniform or random sequence of tones is generated. This acts as the starting point (initial microstate) for the optimisation.
- **Iterative Optimization:** In every iteration one note at a time is modified, and the entire sequence is re-evaluated to calculate the actual macro-properties.
  1. Trial notes are selected, and a tonal centre is computed based on the previous notes, which is used to evaluate the kinetic energy if needed.
  2. The difference between actual and target properties is computed.
  3. The trial notes that reduce this difference are accepted.
  4. This repeats until the deviation from the prescribed properties is minimised.

The algorithm will always yield the same output for any given input prescription, i.e., a deterministic behaviour. It is important to note that the algorithm does not define musical scales, keys, rhythms, or chord progressions. As demonstrated in the results section, these musical features emerge naturally for specific landscapes of the entropy-energy domain between the ground state and maximum entropy (equilibrium). This setup also allows real-time translation of instantaneous prescriptions to tones, so that the system functions as a 'smart instrument' that adapts real-time emotional or gestural input and returns tonal sequences matching the desired expressive profile. An Android implementation of the generative system, DJ Bach, is available on Google Play. The app can have three input sources for E and S prescriptions, including EEG features from the Muse brain-sensing headband from Interaxon Inc., Canada.

## 3 Results

This section presents results from applying the *Tonamic* system to two use cases: the analysis of existing melodic lines using the entropy-energy formalism, and the generation of original music based on prescribed affective trajectories. The outcomes confirm that the model can show salient features in traditional melodies and generate perceptually coherent music without using predefined musical rules or training data.

### 3.1 Analysis of Traditional Melodies and Compositional Rules

To demonstrate the analytical capacity of the framework, two well-known melodies were examined (Fig. 1). Electronic renditions of the analysed excerpts are available in Refs [20, 21]. Both were analysed using a fixed moving window to track variations in energy (blue, melodic dissonance), entropy (green, unpredictability) and their ratio (red, expressive tension). In both melodies, the method identifies natural tension–release structures in the energy trajectory on both micro and macro scales. For the second melody, however, a gradual buildup of dissonance followed by resolution on the macro scale is much more prominent. A more focused analysis was performed on a traditional/historical melodic exercise in tonal composition: outlining the augmented fourth in a *Cantus Firmus* line [22]. Two note sequences were analysed, one adhering to conventional guidelines (F-A-G-B-A, "OK") and one violating them (G-F-G-B-A, "Not OK"). The two sequences differ only in the ordering of notes, making them ideal for testing sensitivity to subtle structural changes. The "Not OK" version shows a sharp increase in energy towards the end of the phrase, indicating unresolved tension, whereas the "OK" version displays a bell-shaped curve in the energy profile, with tension gradually building and then resolving, consistent with established principles of musical phrasing and closure.

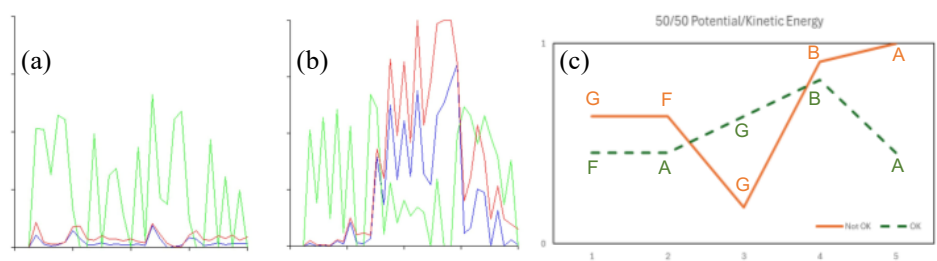

**Fig. 1.** Variation of the macro-properties in a progressively expanding frame for **(a)** the theme of Choral, Symphony No. 9, Beethoven [20], **(b)** Thema Regium, The Musical Offering, J.S. Bach [21], and **(c)** two *Cantus Firmus* examples ("OK" and "Not OK). In (a) and (b) all three properties (energy, entropy and their ratio) are shown, whereas only energy is shown in (c). The horizontal axis is the number of tokenised events in all plots and the vertical axis is the normalised property between 0 and 1.

### 3.2 Generation of Emotionally Structured Music

In the generative mode, the user prescribes dynamic variations of energy and entropy over time. The system then constructs sequences of discrete tones whose properties match the desired shape. Several examples of the electronic rendition of the output can be found on soundcloud.com/tonamic. Notable examples of output include Ref. [23], which resulted from a simple sinusoidal energy variation with three minima (E=0) at the beginning, after the midpoint, and at the end. Optimisation in this example is performed only with respect to E, thus, resulting in rhythm obscurity. A notable example of output with rhythmic clarity is given in Ref. [24]. This was obtained by taking two simple sinusoidal variations of E and S as input, with different wavelengths and amplitudes. This early output of the system demonstrates rhythmic emergence though with frequent modulations and obscure tonal centre. Despite the lack of a predefined scale of rhythm, both pieces exhibit emergent melodic direction and periodicity, indicating that the prescribed trajectories of E and S alone can yield music-like output.

# 4 Conclusion

This study presented a generative and analytical framework, *Tonamic*, for music based on thermodynamically inspired macro-properties, namely entropy and energy, without relying on musical rules, training data, or predefined stylistic structures. By treating tones as microscopic elements and measuring their combined behaviour over time, the system allows for both the analysis of existing melodies and the generation of new music with emotionally meaningful structure. The system can be used for music analysis and classification, besides serving as a tool for composition and adaptive music generation in real-time, e.g., for neurofeedback or gaming applications. Future directions can include exploring tuning systems other than the 12-note equal-tempered scale, music without a traditional tonal centre, and music from non-Western harmonic systems, as well as experiments to assess the correlation between the perceived and the prescribed emotional variables.

# gem5/Z3/gcc/Clang/Redis glibc Heap Fitness Landscapes

William B. Langdon, Justyna Petke, David Clark

W.Langdon@cs.ucl.ac.uk j.petke@ucl.ac.uk david.clark@ucl.ac.uk
CREST, Department of Computer Science,
UCL, Gower Street, London, WC1E 6BT, UK

**Abstract.** We adapt "The gem5 C++ glibc Heap Fitness Landscape" W.B. Langdon and B.R. Bruce GI@ICSE 2025, to use Valgrind Massif on 1 300 000 line C++ gem5, on 600 000 LOC C++ theorem prover Z3 and benchmarks from SMT-COMP 2024. Showing the memory landscape is far smoother than is commonly assumed and that Magpie and CMA-ES can tune GNU malloc giving 2.4 megabytes reductions in peak RAM use without coding changes. Similar results are given on the GCC and Clang LLVM compilers and 150 000 LOC C Redis key-value database.

**Keywords** SBSE, computer program tuning, parameter search

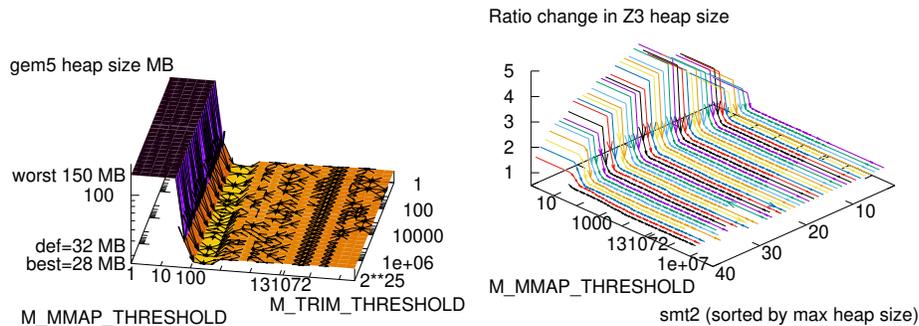

**Left:** Plot of 2 of GNU glibc malloc's 7 parameters. Setting mmap infeasibly small increases the heap size (blue-black), but for gem5 values 256–512 give a reduction 11% (bright yellow) relative to the default 131 072 (orange). Of the 7 dimensions, one (mmap) is dominant and it has a huge sweet spot. **Right:** Z3 with 40 SMT problems varying only mmap. Again arrows show direction of improvement. Note log scales.

## 1 Summary

We re-apply genetic improvement [9,6,12,4,11,1,3], to optimise gem5's heap [8] using the newest version of Magpie [2]. Additionally we systematically explore the glibc heap optimisation search landscape of: Microsoft's theorem prover (Z3 [10] 600 000 lines of C++), the GCC and LLVM Clang C++ compilers and the 150 000 LOC C++ Redis Ltd. key-value store database, including also experiments with both Magpie and CMA-ES [5]. Whilst the latter four landscapes are, like gem5, also smooth, they are disappointingly flat and Magpie, by searching for longer, does better than CMA-ES.

## 2   Introduction

It is sometimes assumed that the fitness landscapes are vast and very difficult. In [8] we showed for a million C++ program, gem5, that this need not be the case and that sometimes real programs give a landscape which is smooth, effectively unimodal and contains many acceptable solutions (left Figure 1). The right hand side of Figure 1 shows this can be true of Z3.

## 3   glibc Heap 3 Important Parameters

There are 7 GNU malloc tuning parameters. For simplicity we avoid multi-threading, which means only 3 are relevant (see Table 1). Here, like gem5, MMAP_THRESHOLD is dominant.

## 4   Massif Measuring the Maximum Heap Size

Valgrind's performance tuning tool Massif, can report both dynamic and peak heap usage with high precision. However Massif imposes, particularly if very accurate results are wanted, a runtime overhead. For example, Valgrind 3.23.0 Massif (`--pages-as-heap=yes --peak-inaccuracy=0.0 --time-unit=B`) slowed down gem5 by about 15 fold, Z3 by 5, Clang by 6, and Redis by 4 fold. (On the other hand it had almost no impact on gcc.) In our earlier work [8] we used the GNU malloc_info function, which imposes little overhead. malloc_info requires minor code changes but reports the current heap state, rather than the peak, and so care is needed as to exactly when it is called.

### 4.1   Massif Fitness Function

The four new benchmarks (Z3, gcc, Clang, Redis) essentially use the same fitness function. The program is run with Massif and the default parameters and then it is run again with Massif and the mutated values of MMAP_MAX, MMAP_THRESHOLD and TRIM_THRESHOLD. Fitness is the ratio of the peak heap usage. If any of the program's outputs are different, the mutant is abandoned.

## 5   Examples: gem5, Z3, gcc, Clang, Redis

**gem5**  We described the gem5 benchmark in [8]. The mean distribution of the geometric distribution was corrected to be the same as the glibc defaults. (This was intended, but previously, due to a misunderstanding, it was $2.56\times$ bigger [8].)

**Table 1.**  New Magpie parameter file `mallopt30.txt` E.g. 1st `g[][]` indicates mutations of parameter `MMAP_MAX` are drawn from a geometric distribution, bounded by 0 to 33554432, mean 65536, and default (unmutated) value of 65536.

```
TIMING="run"
M_MMAP_MAX_tune         g[0,33554432,1/65536][65536]
M_TRIM_THRESHOLD_tune   g[0,33554432,1/131072][131072]
M_MMAP_THRESHOLD_tune   g[0,33554432,1/131072][131072]
CLI_PREFIX = ""
CLI_GLUE = "="
```

**Z3** The annual international SMT-COMP competition pits SMT solvers against each other on many benchmarks. We choose Certora Prover's QF_UFDTLIA [7], as last year (SMT 2024) Microsoft's Z3 [10][1] did pretty well on it[2]. QF_UFDTLIA contains 76 .smt2 files. We chose the 16th as the best compromise between challenging Z3 and runtime. Like gem5, gcc and Redis, Z3's peak heap memory usage is at the end Figure 2. (Clang's peak heap occurs about 80%.)

**gcc** We had previously downloaded MySQL[3]. We selected its largest human written C++ source file, ha_innodb.cc, 15 000 LOC. The fitness function compiles it with gcc version 13.3.1 using the same command line and include files as were used to build MySQL (with include files 210 000 lines). To simplify comparing binary object files with and without Massif, we added `-frandom-seed`.

**Clang** We compiled the same files as gcc with Clang 18.1.8

**Redis Ltd.** [4]. `redis-server` holds the key-value store. Massif is used to report its heap, whilst `redis-benchmark` exercises it. Thus the fitness function simultaneously runs `redis-server` and `redis-benchmark`. To avoid variability, it was run with a fixed seed, a large (10 000 bytes) SET/GET value and a single client connection. Using Massif limited the number of requests (1024-32 = 992). `redis-benchmark -P 16 -p $PORT -c 1 -d 10000 -n 992`.

## 6   Search Tools: Grid search, Magpie and CMA-ES

**gem5 Two Dimensional Grid Search** Work on gem5 [8] showed GCC malloc parameter MMAP_MAX, as long as a reasonable value is used, has little impact. Hence in our grid searches it was left at its default value (Table 1). The other two parameters have default values of $2^{17}$. In [8] the two dimensional grid search varied both covering all integer and half integer powers of two from 1 to $2^{25}$ Figure 1 (left). Arrows towards fitness improvements emphasis that although the gem5 fitness landscape is smooth, it is not flat.

**One Dimension Grid Search: Z3, gcc, Clang, Redis** Figure 1 (left) emphasis that TRIM_THRESHOLD (like MMAP_MAX) has relatively little impact and so for the one dimensional grid searches it was left at its default value and only MMAP_THRESHOLD was varied. However the range was increased to $2^{35}$.

**Magpie** We use Magpie downloaded on 19 Feb 2025[5].

**CMA-ES** We also use Hansen's CMA-ES[6]algorithm [5].
Both Magpie and CMA-ES were run ten times on Z3, gcc, Clang and Redis.

## 7   Results

With the corrected means to the distribution of Magpie mutations, see Table 1, the number of Magpie gem5 runs which found good solutions increased from 1/10 to 6/10 (median saving 12.0% ± 0.1%, MMAP_THRESHOLD 300 ± 100).

---

[1] Z3 downloaded `https://github.com/Z3Prover/z3/` 4 Feb 2025

[2] `https://zenodo.org/records/11061097/files/QF_UFDTLIA.tar.zst` 4 Feb 2025.

[3] `https://github.com/mysql/mysql-server` 12 Mar 2025

[4] `https://github.com/redis/redis/archive/refs/heads/unstable.zip` on 1 Apr

[5] `https://github.com/bloa/magpie`

[6] `https://cma-es.github.io/`

All ten Magpie and all ten CMA-ES runs found solutions which reduced Z3's peak heap usage by 1.5% (Magpie median MMAP_THRESHOLD $30\,000 \pm 600$ and CMA-ES median $50\,000 \pm 10\,000$.) See also purple line in Figure 3. Figure 3 $+/\times$ shows heap improvement/worsening varies smoothly with MMAP_THRESHOLD for Z3, gcc, Clang and Redis. The mean percentage best improvement found by ten Magpie and ten CMA-ES (both with up to 1000 fitness trials) are shown in Table 2.

**Table 2.** Mean percentage improvement of ten runs of Magpie and CMA-ES

|        | Z3           | (best) | gcc         | (best) | Clang       | (best) | Redis       | (best) |
|--------|--------------|--------|-------------|--------|-------------|--------|-------------|--------|
| Magpie | $1.5 \pm 0.0$ | 1.5    | $0.0 \pm 0.0$ | 0.0    | $0.0 \pm 0.0$ | 0.1    | $0.3 \pm 0.0$ | 0.3    |
| CMA-ES | $1.5 \pm 0.0$ | 1.5    | $0.0 \pm 0.0$ | 0.0    | $0.0 \pm 0.0$ | 0.0    | $0.2 \pm 0.1$ | 0.3    |

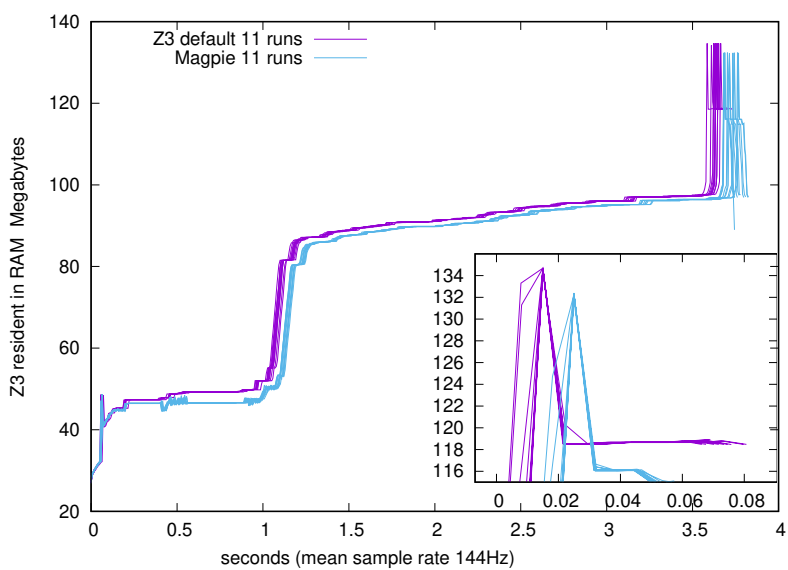

**Fig. 2.** Z3 use of heap. Magpie chose MMAP_THRESHOLD=29 918. The lower trace shows it saves 2.4MB early in the run but is slightly slower. In the insert the time axis is shifted so the plots synced at the peak value.

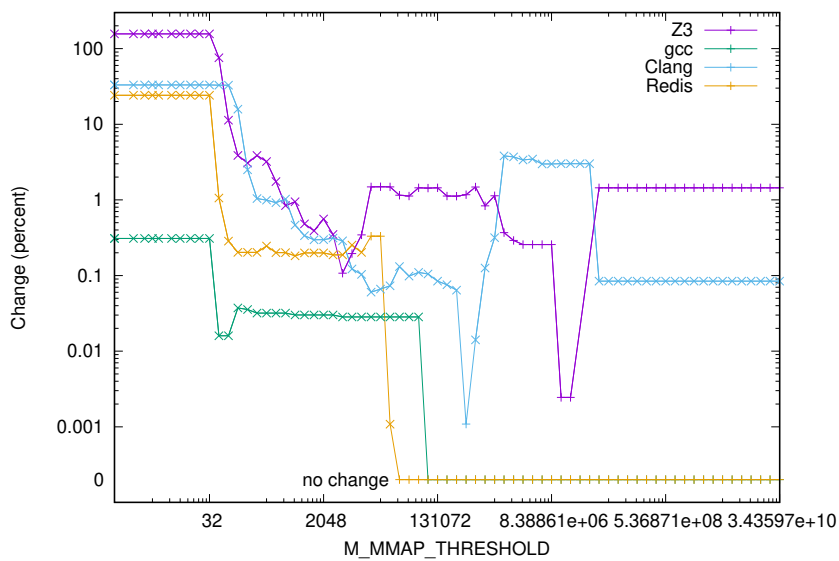

**Fig. 3.** × percentage increase in peak heap. + percentage reduction. Notice log scales exagerate small differences.

# Multi-Modal Fusion Techniques for Detecting Abnormal Events in Videos

Ta Tran Thanh Tung, Dinh Cong Binh, Vu Minh Hoang, Vuong Tu Binh, and Tran Minh Hoan

Department of Artificial Intelligence, FPT University, Hanoi, Vietnam

**Abstract.** Abnormal event detection in videos is critical for public safety, especially with large-scale surveillance. While deep learning improves accuracy, existing methods struggle to fuse multimodal data like RGB, audio, pose, and optical flow due to heterogeneity and misalignment. We propose a cross-modal attention and gating network to enhance feature fusion. Our method introduces Modality-wise Feature Matching Subspace (MFMS) and Cross-Modal Attention (CMA) to improve representation consistency. Experiments on the XD-Violence dataset achieve an average precision of **84.98%**, confirming the method's effectiveness.

**Keywords:** Multimodal anomaly detection · Cross-modal attention · Feature alignment · Video surveillance

## 1 Introduction

Anomaly detection in videos involves diverse modalities such as RGB, optical flow, pose, and audio [2]. While unimodal methods use one data type, multimodal fusion captures complementary cues. Existing methods suffer from ineffective feature alignment, noise, and lack of adaptive fusion [9,12].

Traditional handcrafted approaches [7] lack scalability. Deep learning models like CNNs and transformers [5] improve representation, while hybrid techniques combine strengths but still face fusion issues [4].

We propose a fusion framework that integrates RGB, flow, audio, and pose by aligning features in a shared space. A Modality-wise Feature Matching Subspace (MFMS) reduces inconsistency, while Gated Cross-Modal Attention dynamically weights modalities [3]. Using I3D [1] and VGGish [6], our model achieves 84.98% AP on XD-Violence.

## 2 Method

Our method includes three stages: feature extraction, alignment, and gated fusion. RGB and flow features are extracted with I3D, audio with VGGish, and pose with PoseC3D. MFMS aligns all modalities in a shared latent space. Self- and cross-attention refine the alignment. Gated attention adaptively fuses them into a unified feature. Anomaly scoring is performed with a regression head.

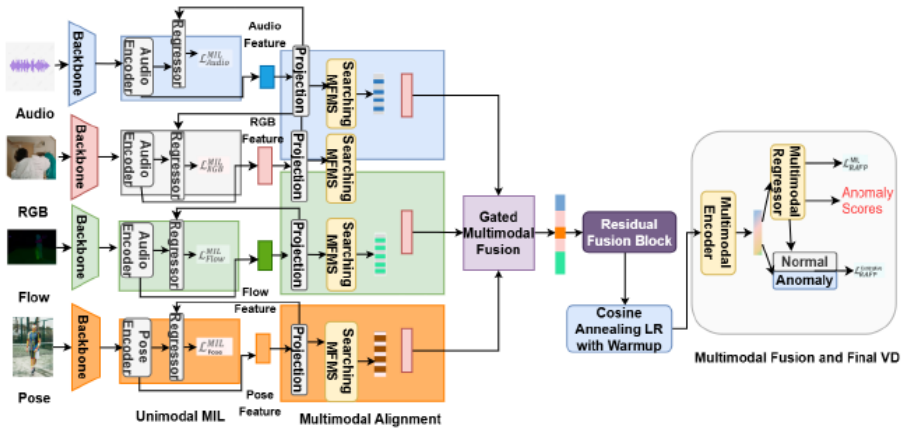

**Fig. 1.** Overview of the proposed multimodal anomaly detection framework.

## 3 Experiments

### 3.1 Dataset and Metrics

We evaluate on XD-Violence, a benchmark with over 200 hours of multimodal videos. Frame-level Average Precision (AP) and Area Under Curve (AUC) are used.

### 3.2 Results

Our model achieves **84.98% AP**, outperforming state-of-the-art methods like MSBT (84.32%). Including pose improves performance, proving its complementary value.

Table 1. Performance Comparison on the XD-Violence Dataset

| Method | Modality | AP (%) |
|---|---|---|
| SVM baseline | RGB | 50.78 |
| OCSVM [8] | RGB | 27.25 |
| Hasan et al. [5] | RGB | 30.77 |
| Sultani et al. [9] | RGB | 75.68 |
| RTFM [11] | RGB | 77.81 |
| HL-Net [12] | RGB + Audio | 78.64 |
| ACF [12] | RGB + Audio | 80.13 |
| Zhang et al. [12] | RGB + Audio | 81.43 |
| MSBT [10] | RGB + Audio + Flow | 84.32 |
| **Ours** | RGB + Audio + Flow | 83.51 |
| **Ours** | RGB + Audio + Flow + Pose | 84.98 |

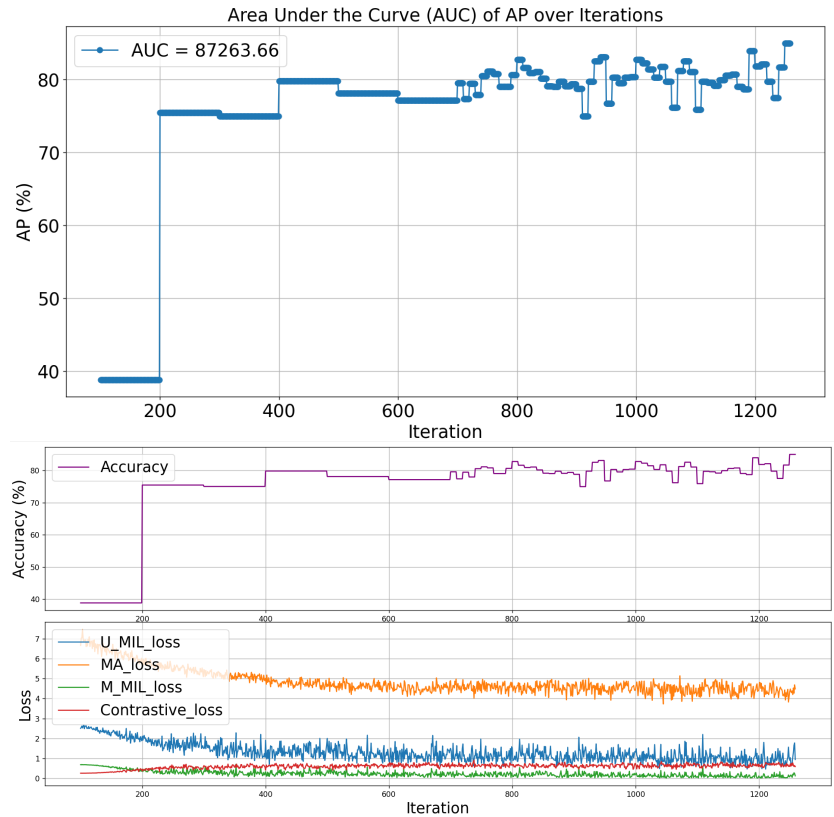

**Fig. 2.** (Top) AUC over iterations. (Bottom) Training loss and accuracy trends.

## 4  Conclusion

We introduce a robust fusion framework using MFMS and gated attention to align and integrate RGB, flow, audio, and pose for anomaly detection. Our approach achieves state-of-the-art results on XD-Violence.

# Testing Self-Organized Load Balancing in Distributed Systems

Juan Luis Jiménez Laredo[1][0000−0002−9416−2005], Juan Julián Merelo
Guervós[1][0000−0002−1385−9741], Paulin Héleine[2], and Damien
Olivier[2][0000−0002−6552−8151]

[1] TIC024-ICAR, Universidad de Granada, Spain
[2] RI2C-LITIS, Université Le Havre Normandie, France

**Abstract.** This paper investigates the dynamic load balancing capabilities of a sandpile-based heuristic under a ramp-up workload scenario. By simulating a gradual increase in task arrivals, emergent self-organizing behavior efficiently redistributes loads across processing elements (PEs) while minimizing energy usage. Experimental results demonstrate that the heuristic dynamically recruits resources in a near-optimal fashion, with energy consumption closely tracking the growing demand.

**Keywords:** Dynamic load balancing · Energy efficiency · Sandpile · Ramp-up workload · Self-organized criticality

## 1 Introduction

Distributed systems require efficient load balancing mechanisms to maintain performance and reduce energy consumption. Traditional optimization techniques, such as evolutionary meta-heuristics [3, 2], often suffer from high overhead and lack real-time adaptability. In contrast, our approach leverages self-organized criticality as observed in sandpile models [1], where simple local rules lead to global load distribution.

In our model, tasks are akin to sand grains while processing elements (PEs) function as grid cells in the classical Bak-Tang-Wiesenfeld (BTW) sandpile model. When a PE's task count exceeds a critical threshold, the model triggers an avalanche—a local redistribution of tasks to neighboring PEs. This not only balances the load but also minimizes energy consumption since idle PEs (assumed to have zero energy use) are only activated when needed. Recent work in energy-aware resource management [7, 6] further motivates this design.

## 2 Model Description

The model adapts the BTW sandpile to a distributed computing environment as follows. Each PE, representing a grid site, collects tasks (grains). When the number of tasks $h(p)$ at a PE $p$ reaches a critical threshold $h_c$, an avalanche occurs, transferring a fixed number $m$ of tasks to each neighboring PE in a von

Neumann neighborhood. With standard parameters $h_c = 4$, $m = 1$, and range $r = 1$, the stability constraint is:

$$h_c \geq m \times 2r(r + 1).$$

Algorithm 1 summarizes the decentralized load balancing process. Each overloaded PE autonomously redistributes tasks to its neighbors, while idle PEs process tasks without incurring energy costs.

---

**Algorithm 1** Sandpile Dynamic Load Balancing

---

1: **while** there exists a PE $p$ with $h(p) \geq h_c$ **do**
2:     **for all** neighbors $p'$ of $p$ **do**
3:         Transfer $m$ tasks from $p$ to $p'$
4:     **end for**
5: **end while**
6: **for all** idle PEs $p$ with $h(p) > 0$ **do**
7:     Process the next task in the queue
8: **end for**

---

## 3   Experimental Evaluation: Ramp-Up Scenario

We simulate a ramp-up scenario in which tasks are progressively injected into the system at an increasing rate. This controlled workload variation allows us to study how the sandpile-based heuristic adapts resource allocation and energy consumption dynamically.

Figure 1 captures the essential system dynamics:

- **Workload Profile:** Subfigure 1a shows the gradual increase in tasks over simulation cycles.
- **Resource Activation and Energy:** Subfigure 1b depicts the corresponding recruitment of processing elements and the energy consumed (with active PEs incurring 1 energy unit per cycle).

The energy consumption curve closely mirrors the increase in active resources, indicating that only the necessary number of PEs is activated as the load intensifies.

### 3.1   Discussion

The ramp-up experiment reveals several key insights:

- **Adaptive Resource Recruitment:** As the task rate increases, more PEs are activated in close relation to the load. This confirms the effectiveness of local task redistribution—the avalanche mechanism—and that resources are not wasted.

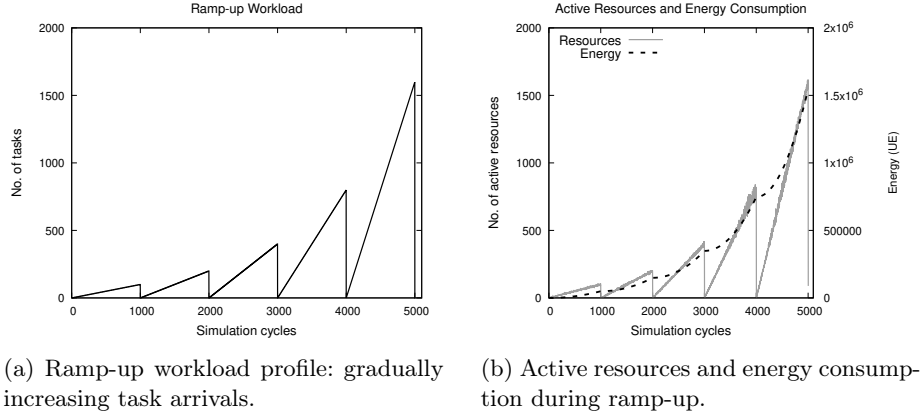

(a) Ramp-up workload profile: gradually increasing task arrivals.

(b) Active resources and energy consumption during ramp-up.

Fig. 1: Emergent behavior under a ramp-up workload. The left subfigure illustrates the growing workload, while the right subfigure shows the corresponding resource activation and energy usage.

- **Energy Efficiency:** By ensuring that idle PEs consume no energy, the system's energy usage remains nearly optimal. The close correspondence between the active resource and energy curves supports claims made in energy-aware scheduling [7, 6].
- **Scalability and Robustness:** The simplicity of the sandpile rule allows the system to scale with minimal computational overhead compared to traditional methods [3, 2]. The emergent global stability achieved through local interactions also suggests robust performance under varying load conditions.

## 4    Related Work and Extensions

Previous studies have explored sandpile-inspired methods for load balancing in distributed systems [4, 5]. These methods have demonstrated that emergent self-organization can result in efficient scheduling and energy savings in real-world settings. Our work extends these ideas by focusing on a synthetic ramp-up workload, thereby providing clear insights into the dynamic adaptation properties of the heuristic.

Future work could explore hybrid scheduling strategies, where the initial allocation is managed by methods such as round-robin, with the sandpile mechanism triggered for local rebalancing. This combination could further reduce response times and enhance energy efficiency.

## 5    Conclusions

In this paper, we evaluated a sandpile-based heuristic for dynamic load balancing under a ramp-up workload scenario. The experimental results illustrate that the model dynamically recruits processing elements as needed, with energy consumption closely following the active resource profile. Compared with traditional high-overhead methods [3, 2], the proposed decentralized approach demonstrates substantial promise for real-time, energy-efficient scheduling in distributed systems.

Future research will focus on integrating this heuristic with complementary scheduling techniques and validating the approach through more extensive simulations and potential real-world implementations.

## Acknowledgements

This work was supported by the Plan Propio de Investigación y Transferencia project PPJIA2023-031, Proyecto Plan Operativo FEDER Andalucía project C-ING-027-UGR23 and the Ministerio Español de Ciencia e Innovación under project numbers PID2023-147409NB-C21 and PID2020-115570GB-C22.

# Multi-objective particle swarm optimization for environmental risk/benefit analysis

Luca Puzzoli[1], Gabriele Sbaiz[2][0000−0003−3500−6843], and Luca Manzoni[1][0000−0001−6312−7728]

[1] Department of Mathematics, Informatics and Geosciences, University of Trieste, Via Valerio 12/1, 34127 Trieste, Italy
luca.puzzoli@studenti.units.it, lmanzoni@units.it
[2] Department of Economics, Business, Mathematics and Statistics, University of Trieste, Via A. Valerio 4/1, 34127 Trieste, Italy
gabriele.sbaiz@deams.units.it

**Abstract.** Hydropower is a fundamental renewable energy source, and the Amazon basin represents one of its largest untapped frontiers. However, its expansion in this ecologically sensitive region raises significant environmental challenges, especially concerning greenhouse gas emissions. In this paper, we develop a multi-objective optimization framework that employs a variant of the Multi-Objective Particle Swarm Optimizer to balance the competing objectives represented by the total electricity generation and the reduction of carbon emissions. We analyse a dataset of 509 dams, categorized by geographical and technical features, to assess the impact of site selection. We further inspect the key features of dams that compose the best configurations to maximize energy output while minimizing emissions. In such configurations, the dams are located in highland areas, offering flexible trade-offs and allowing planners to balance sustainability with energy demands. Decision-makers could take advantage of this work by adopting a strategic approach to hydropower expansion that prioritizes energy efficiency and environmental responsibility, showcasing the effectiveness of computational optimization in sustainable energy planning.

**Keywords:** Multi-objective optimization problems · Multi-objective PSO · Risk/benefit analysis · Dams strategic plans.

## 1 Introduction

In this work we address the strategic planning of dams for hydropower development in the Amazon basin, with the dual objectives of maximizing electricity generation and minimizing greenhouse gas (GHG) emissions. If hydropower expansion proceeds without strategic planning, particularly in lowland areas that require large reservoirs, it could lead to substantial GHG emissions, due to the decomposition of flooded organic material. These emissions reduce the climate benefits of hydropower and contribute to global warming. Conversely, a carefully coordinated approach that prioritizes dam placement in higher elevation areas with smaller reservoirs and higher power densities could significantly reduce emissions while still ensuring renewable energy generation. The urgency of addressing this issue is driven by the need to balance electricity generation with the necessity to minimize GHG emissions. This is particularly critical in the context of global climate mitigation efforts, as outlined in the IPCC Special Report [7]. The goal is to optimize dam locations and configurations to maximize electricity generation while minimizing environmental impacts. This necessitates the use of a multi-objective optimization framework to evaluate trade-offs between energy production and carbon emissions. To tackle this problem we employ a variant of the Multi-Objective Particle Swarm Optimization (MOPSO) algorithm. Extending PSO to multi-objective optimization problems was initially done in [5], where it has been shown that swarm-based approaches can handle conflicting objectives effectively, laying the foundation for later adaptations. Later, in [2] the authors have introduced a MOPSO algorithm incorporating Pareto dominance and an external archive to maintain solution diversity and improve convergence. In [4] the authors have investigated the impact of key parameters on MOPSO performance, showing that the inertia weight plays a crucial role in balancing exploration and exploitation. Finally, in [6], the authors have provided a detailed review of MOPSO advancements, categorizing modifications based on leader selection, archive management and velocity updates. The paper is structured as follows: in the next section we introduce the mathematical model and we describe the developed variant of the MOPSO. In Section 3 we present the results and Section 4 concludes.

## 2    Mathematical model and solver

This study presents a multi-objective optimization problem that seeks to balance two conflicting objectives. More specifically,

$$\max_{x \in [0,1]^n} \Big( f_1(x), -f_2(x) \Big) \text{ where } f_1(x) = \sum_{i \in S(x)} \text{EG}_i, \text{ and } f_2(x) = \sum_{i \in S(x)} \text{CI}_i,$$

subject to $S(x) = \{i \mid x_i > 0.5\}$ and $x_i \in [0,1], \forall i \in \{1, \dots, n\}$. Each potential solution is expressed by a decision vector $x = (x_1, \dots, x_n) \in [0,1]^n$, representing the selection probability of each dam, based on a predefined threshold. The first objective function aims to maximize the total electricity generation (EG) from the selected dams. This function sums the electricity generation contributions from all selected dams, aiming to maximize the total energy output. The second objective function, instead, seeks to minimize the total carbon intensity (CI) associated with the selected energy sources. Since these objectives are conflicting, to solve this multi-objective optimization problem, we employ the MOPSO solver which is a PSO variant able to produce the Pareto front efficiently. After some comparison results with different variants of MOPSO, we have chosen for our analysis the MOPSO-based algorithm where we introduce a linearly decreasing inertia weight to encourage an adaptive balance between exploration and exploitation. The inertia weight starts at 0.9 and gradually decreases to 0.4 as the iterations progress. In this model, the social and cognitive coefficients are kept constant at 1.49 to isolate the impact of inertia weight variation (see [3]). The population size is set to 1,000 particles, and the number of iterations is fixed at 400. Finally, the archive size is set to 300 particles and the MOPSO is executed 30 times independently, with each run initialized using a different randomly generated swarm.

## 3    Experimental analysis

We consider a dataset in which there are accounted a total of 509 dams, categorized in 351 proposed and 158 existing. These data are provided in [1], accordingly with information from national government databases for countries, where updated inventory data are readily available. In details, for both proposed and existing dams, the dataset comprehends geographic location, country and region (upland/lowland), elevation and technical data including installed capacity. The alluvial diagram, reported in Figure 1 (right panel), represents the evolution of dam selections across multiple Pareto optimal solutions of a selected run computed by MOPSO (left panel). Each node in the diagram corresponds to a specific solution along the Pareto front, representing a unique configuration of selected dams. The nodes are positioned sequentially, showing the transition from one solution to the next. By analysing these transitions, we can assess whether the Pareto optimal solutions exhibit stability, where the same dams are consistently selected, or variation, where significant modifications occur in dam selection. The presence of wide flows in Figure 1 for the "Selected" dams suggests a core set of dams that frequently appear across multiple solutions, indicating their essential role in balancing electricity production and carbon intensity. Analogously, wide flows for the "Discarded" dams indicate dams that are either never selected among the solutions or were introduced but later removed, suggesting an unstable selection pattern. Therefore, these dams do not consistently contribute to optimal trade-offs and are instead subject to repeated inclusion and exclusion as the Pareto front evolves. In general, the previous results show that the core of the efficient dams (in terms of energy production and climate impact) is solid as well as the pool of the inefficient ones. This implies a well-founded starting point for the discussion of decision-makers in order to choose the dams to be built to adhere to increased demand for energy or sustainability.

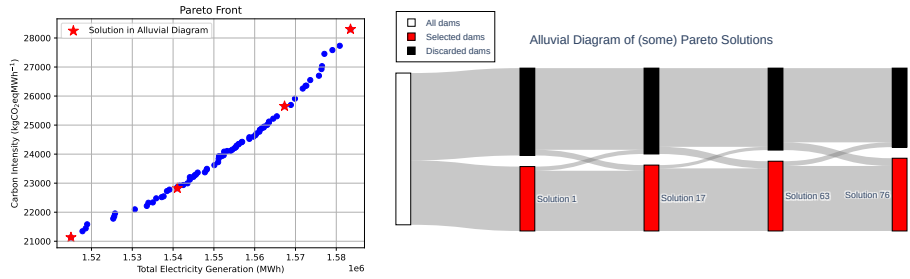

**Fig. 1.** Pareto front and alluvial diagram for selected solutions.

## 4   Conclusions

This study has addressed the strategic planning of hydropower expansion in the Amazon basin. To tackle this challenge, we have developed a multi-objective optimization framework that explores the trade-offs between energy production and carbon emissions. Using MOPSO, we have identified Pareto optimal dam configurations that balance these competing objectives. Additionally, this study has investigated the operational status of existing dams, whether they should remain active or be turned off, and the future potential of proposed dams, evaluating whether they should be constructed, or activated, or not. The findings could contribute to the broader discussion on renewable energy planning, highlighting the importance of strategic site selection in reducing environmental impacts while maximizing energy efficiency.

**Acknowledgments.** The research of G. Sbaiz was partially supported by the project of the Regional Programme (PR) FSE+ 2021/2027 of the Friuli Venezia Giulia Autonomous Region - PPO 2023 - Specific Programme 22/23.

# An Automated Financial Management System for Risk Budgeting Portfolio Optimization

Massimiliano Kaucic[1,2,3[0000−0002−6565−0771]], Filippo
Piccotto[1,2[0000−0002−3869−1471]], and Renato Pelessoni[1,2[0000−0002−7820−8173]]

[1] University of Trieste, Department of Economic, Business, Mathematical and
Statistical Sciences
[2] SOFI Lab, Soft Computing Laboratory for Finance and Insurance
[3] YourValue Finance S.R.L.

**Abstract.** We introduce a novel automated decision support system
for portfolio optimization that maximizes a financial performance mea-
sure subject to cardinality, box, budget, and a set of risk budgeting
constraints. First, we analyze the capability of the developed solver to
identify feasible solutions. Then, we compare the proposed investment
strategy to several common benchmark strategies to assess its profitabil-
ity. The results show that our solver moves efficiently within the feasible
region. Moreover, the risk budgeting-based model attains better ex-post
financial performance compared to the equally weighted portfolio bench-
mark.

**Keywords:** Knowledge-Based Financial Management System; Portfolio Opti-
mization; Risk-Budgeting; DISH-XX; TODIM.

## 1 Introduction

The portfolio selection process typically involves two stages. The first phase
comprises the selection of the most promising stocks to be included in the opti-
mization, while the second concerns the optimal wealth allocation between the
portfolio constituents. This paper presents a novel automated knowledge-based
financial management system built upon two interconnecting modules tailored
for solving portfolio optimization models that maximize a financial performance
measure subject to real-world constraints, namely cardinality, box, budget, and
a set of risk budgeting constraints to provide an explicit control of risk. The first
module employs the multi-criteria decision-making procedure called TODIM [3]
to develop a ranking between stocks concerning several financial criteria iden-
tified by the end user. The first criterion is based on mutual information, and
it is employed to capture the microstructure of the stock market. The second
one is the momentum, and the third is the upside-to-downside beta ratio. With
this step, we can bypass the explicit management of the cardinality constraint
within the optimization process. Next, in the second module, we determine the
optimal portfolio weights using a version of the linear population size reduc-
tion success-based differential evolution algorithm with double crossover (the so-called DISH-XX [4]). This algorithm is extended with an ensemble of ad-hoc constraint-handling techniques for solving the proposed optimization model. The synergy between these two modules can suit the necessities of various types of end users who want to be directly involved in the design of the portfolio strategy.

An extensive experimental analysis is conducted considering the STOXX Europe 600, one of the most significant stock market indices. In the empirical part, we first validate the algorithm's ability to generate feasible solutions, then we point out the profitability of our investment strategy compared to the equally weighted portfolio benchmark.

## 2   The Portfolio Optimization Model

We consider a frictionless market that does not allow for short selling, and the investable universe consists of $n \geq 2$ risky assets. A portfolio is denoted by the vector of weights $\mathbf{w} = (w_1, \ldots, w_n)^\top \in \mathbb{R}^n$, where $w_i$ represents the proportion of capital invested in asset $i$. Let $\mu_p(\mathbf{w})$ and $\sigma_p(\mathbf{w})$ be the expected rate of return of portfolio $\mathbf{w}$ and its volatility, respectively. Then, we tackle the following portfolio optimization problem:

$$\underset{\mathbf{w} \in \mathbb{R}^n}{\text{maximize}} \quad \frac{\mu_p(\mathbf{w})}{\sigma_p(\mathbf{w})^{\text{sign}(\mu_p(\mathbf{w}))}}$$

$$\text{s.t.} \quad \sum_{i=1}^n w_i = 1$$

$$\sum_{i=1}^n \delta_i = K$$

$$\delta_i \, lb_i \leq \delta_i w_i \leq \delta_i \, ub_i, \quad \forall i$$

$$(1 - \nu) \frac{\sigma_p(\mathbf{w} \odot \delta)}{n} \leq \delta_i \, RC_i \leq (1 + \nu) \frac{\sigma_p(\mathbf{w} \odot \delta)}{n}, \quad \forall i$$

with $0 \leq \nu < 1$, $\delta_i \, RC_i$ representing the risk contribution of the $i$-th stock to the portfolio risk, and the variable $\delta_i$ being set to 1 if the $i$th stock is included in the portfolio and 0 otherwise.

## 3   The Proposed Financial Management System

The expert system is built upon the following interconnected modules.
*Stock picking.* Instead of directly handling the cardinality constraint in the optimization process, we use a multi-criteria decision analysis technique called TODIM to handle the cardinality constraint. Additionally, we utilize the equal weighting method to define the contribution of each criterion.
*Weights optimization.* We propose a variant of the DISH-XX algorithm, denoted as DISH-XX-$\varepsilon g$, to search for optimal solutions for the reduced portfolio allocation problem. Specifically, we equip the original procedure with the following hybrid constraint-handling technique:

1. To simultaneously fulfill budget and box constraints, we implement the random combination proposed in [1] and the scaling procedure adopted in [2].
2. We address risk budgeting constraints using the $\varepsilon$-constrained method. In particular, we adaptively control the $\varepsilon$ parameter, setting the initial $\varepsilon$-level equal to the mean constraint violation of the best half of the initial population.
3. To exploit the gradient information of the risk budgeting constraints and to move infeasible solutions faster toward the feasible region, we utilize a gradient-based mutation.

## 4   Experimental Analysis and concluding remarks

For the empirical analysis, we consider the daily closing prices of 535 constituents from the STOXX Europe 600 index for the period from December 31, 2014, to October 31, 2024. The case study employs a rolling window investment plan with monthly portfolio rebalancing, with an out-of-sample window covering the period from January 31, 2017, to October 31, 2024.

To evaluate the effectiveness of the solver in finding feasible solutions, a date from the 94 available in the ex-post window is randomly selected, and different portfolio configurations are considered. Figure 1 illustrates the promising results in terms of the evolution of the average values for the feasibility ratio and the population diversity over the generations for different values of $K$ and $\nu$.

To demonstrate the profitability of the proposed investment model, we compare its ex-post results with the equally weighted portfolio constructed using all the assets in the investable universe. The equity lines displayed in Figure 2 highlight the capability of the proposed strategy to efficiently mimic the behavior of the equally weighted benchmark with a very limited number of stocks.

We are currently working on two directions to improve the applicability of the proposed strategy. On the one hand, we are studying mechanisms to implement transaction costs in the rebalancing phase. On the other hand, we are expanding the set of criteria to be included in the TODIM procedure to identify the most performant stocks.

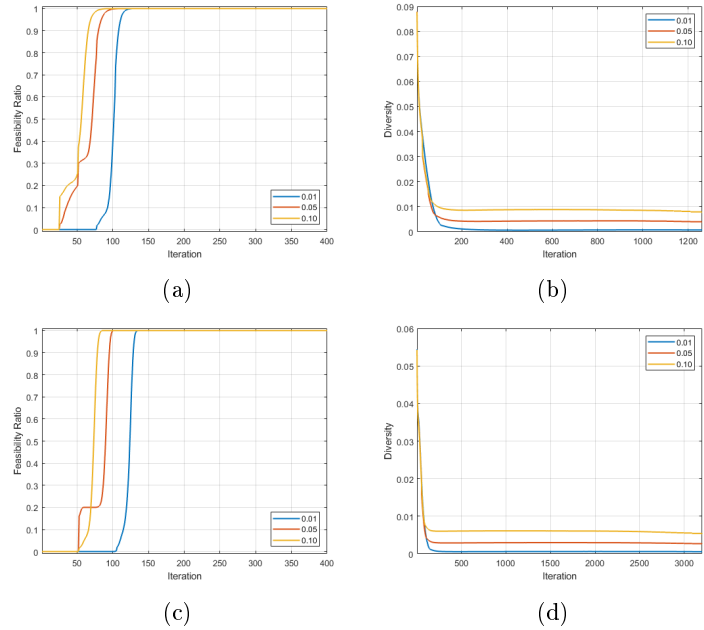

Fig. 1: Average results for 30 random initial portfolio configurations of the feasible ratio and the diversity measure by varying the deviation parameter $\nu$ in $\{0.01, 0.05, 0.10\}$. Plots (a) and (b) show the results for $K = 26$ (5%), plots (c) and (d) show the results for $K = 53$ (10%).

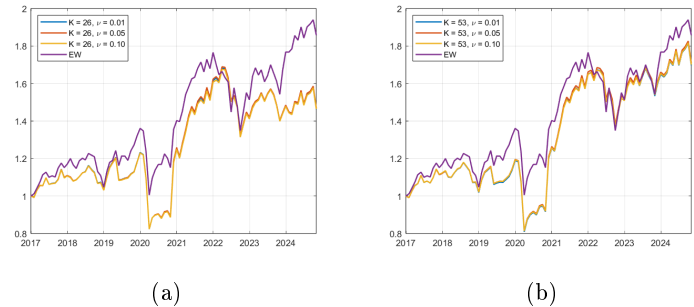

Fig. 2: Wealth evolution of the proposed strategy by varying $\nu$ in $\{0.01, 0.05, 0.10\}$ compared to the benchmark (EW). Plot (a) shows the results for $K = 26$ (5%), and plot (b) shows the results for $K = 53$ (10%).

# Extending Instance Space Analysis via Bipartite Network Communities

Anthony Rasulo[1,2][0009-0008-6664-4235], Julia Handl[2][0000-0002-4338-1806],
Kate Smith-Miles[1][0000-0003-2718-7680], Mario Andrés
Muñoz[1][0000-0002-7254-2808], and Manuel
López-Ibáñez[2][0000-0001-9974-1295]

[1] The University of Melbourne, Australia
{smith-miles,munoz.m,a.rasulo@student.}@unimelb.edu.au
[2] The University of Manchester, United Kingdom
{manuel.lopez-ibanez,julia.handl}@manchester.ac.uk

**Abstract.** We investigate how to obtain human-understandable insights into the joint relationship between algorithm performance, algorithm parameters, and problem instance features. We propose a framework that integrates community detection, from network science, with Instance Space Analysis (ISA), via a bipartite network representation.

**Keywords:** Explainability · Instance Space Analysis · Configuration

## 1 Introduction

Instance Space Analysis (ISA) [6] [7] has been applied in various domains for analysing algorithm strengths and weaknesses via instance features, enhancing test instance diversity, and more. For example, ISA has analysed black-box optimisation algorithms [2] such as CMA-ES variants.

Most algorithms have parameters that affect performance, but ISA does not inherently incorporate algorithm parameter spaces. Different parameter configurations can be treated as different "algorithm implementations", but the number of unique "algorithms" defined makes this naive approach unscalable. Prior work approached this gap via a dual-space projection [5], though existing ISA methods require adaptation for this approach. Instead, we propose integrating community detection from network science with ISA, allowing existing ISA methods to be applied with algorithm parameter spaces naturally integrated.

## 2 Instance Space Analysis via Bipartite Networks

Suppose configuration evaluation metadata has been obtained by evaluating a set $\mathcal{C}$ of algorithm configurations on an instance set $\mathcal{I}$, storing the values of instance features, configuration parameters, and the algorithm performance metric. From this metadata, we represent key relationships via a bipartite network $G$ with partite sets $\mathcal{I}$ and $\mathcal{C}$. For the edge set $E$, edges $\{i, c\}$ represent the performance of algorithm configuration $c \in \mathcal{C}$ on instance $i \in \mathcal{I}$. Two ways to define the edge set are as follows: (1) if configuration $c$ has 'good' performance on instance $i$ then $\{i, c\} \in E$, or (2) if $c$ is evaluated on $i$ then $\{i, c\} \in E$ with the edge weight representing performance. Lastly, the node attributes for $\mathcal{I}$ and $\mathcal{C}$ are instance features and parameters, respectively.

### 2.1   Interpretable Community Detection

In network analysis, community detection aims to group nodes together into densely connected clusters [4] or 'communities', with high intra-cluster connectivity and low inter-cluster connectivity.

In this work, we focus on integrating community structures into ISA, rather than on detection methods themselves. We assume communities are given (found via community detection methods). Grounding this assumption, note that community detection for bipartite networks, in particular, has been studied at depth [1]. In our approach, membership in bipartite communities is defined by simple, interpretable rules in terms of node attributes (via decision trees). The network edge structure (algorithm performance) can thus be related to the node attributes (features and parameters) in an easily interpreted manner. Community detection methods, in this methodology, would aim to identify informative groupings of instances/configurations (defined separately for each partite set), via interpretable binary splits on the features/parameters. Instance communities would then be assigned configuration communities with good expected performance ('co-clusters'). However, in the following, we focus instead on integrating community structures into ISA.

ISA analytical tools (as used for algorithm performance analysis) can instead analyse the performance of configuration communities defined in terms of interpretable rules. The performance metric $y$ instead returns a value $y_{\alpha,x}$ for each configuration community $\alpha$ and instance $x$. Structurally, each $\alpha$ corresponds to an 'algorithm' in ISA. In conjunction with the rules defining the configuration communities, algorithm parameter spaces are naturally integrated within ISA. Moreover, ISA analytical tools, such as footprints (regions in instance space of expected good/best performance) [8], can be directly applied without modification. This approach avoids the scalability issue of otherwise treating each configuration as a distinct "algorithm", by grouping algorithm configurations into communities (obtained in practice via community detection methods).

## 3   Synthetic Experiment

We generate synthetic metadata, idealised to illustrate the proposed methodological extension of Instance Space Analysis (ISA):

1. Sample node attribute values ('features' and 'parameters'), from a uniform distribution $Uniform(0, 1)$ for each dimension of the attribute vectors.

2. Construct the two community trees (decision trees defining interpretable communities) using these values. The trees are iteratively constructed via random splits on a leaf node randomly selected at each stage (but starting with the root node), with the split feature randomly chosen and the split value as the median of the feature (over the assigned elements).
3. Randomly assign co-clusters for instance communities, with replacement.
4. For any instance community $a$ and assigned $b$, in the bipartite network create an edge $\{i, c\}$ for any $i \in a$ and $c \in b$, and there are no other edges.

In the following analysis, instances correspond to nodes in partite set $\mathcal{I}$, with features as the node attributes, and configurations correspond to nodes in partite set $\mathcal{C}$, with parameters as the node attributes. The performance of configuration $c$ on instance $i$ is the weight of edge $\{i, c\}$, or the presence/absence of an edge for a binary performance metric. This follows the same structure as the bipartite network representation of other instance-configuration metadata, and so this functions as synthetic metadata for the proposed extension.

Then, we apply the ISA toolkit [3] to this metadata. Because the configuration communities structurally correspond to 'algorithms' in ISA, the performance of configurations in a community are aggregated to obtain a single performance value for the community on an instance; this is done by averaging, or equivalently in this case, by taking the proportion of edges that exist between instance $i$ and the configuration community. Hence, Figure 1 illustrates an analysis that jointly factors in algorithm performance, algorithm parameters, and instance features.

In the upper left, a configuration community tree shows the defining rules in terms of algorithm parameters, as per the ground truth. The left path is taken if a split condition is true. For example, if parameter $C0 \leq 0.530$ then the configuration is assigned to community CC0 or CC3, depending on whether $C4 \leq 0.477$. In the lower left, feature distributions are shown over the instance space (axes $z_1$ and $z_2$ are linear projections of the features). On the right, the instance space is divided into regions, or footprints, showing where each configuration community is best-performing. In the upper part of the space, configuration community CC3 does the best, and this area has features F0 and F1 high, indicated by the yellow colour in the feature plots and the blue region in the portfolio footprints plot. By the community tree, configuration CC3 is defined by parameter $C0 \leq 0.530$ and parameter $C4 > 0.477$. As such, we can visualise relationships between the aspects of analysis: parameters, features, and performance.

## 4 Conclusion

We have proposed a framework for visual insights into the joint relationship between algorithm performance, algorithm parameters, and instance features; this allows for more direct application of existing ISA methods than previous work [5]. However, the communities were given a priori. In practice, these need to be learned from performance metadata. Therefore, future work will focus on the application and development of community detection methods, tailored to this extension of ISA that naturally incorporates algorithm parameter spaces.

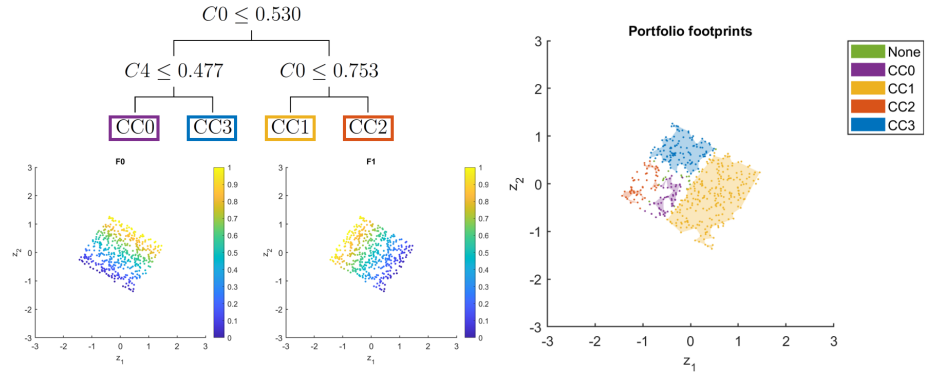

Fig. 1: Configuration communities, footprints (best), and feature plots. C0 and C4 are node attributes of $\mathcal{C}$ (i.e. parameters), and CC0...CC3 are configuration communities partitioning $\mathcal{C}$. F0 and F1 are attributes of $\mathcal{I}$ (i.e. features). The right plot shows which configuration communities are best suited to the regions.

**Acknowledgments.** This research was supported by an Australian Government Research Training Program (RTP) Scholarship, via a dual-award program between University of Melbourne and University of Manchester. This research was (fully/partially) funded by the Australian Government through OPTIMA, Project ID IC200100009.

# Image classification by evolving bytecode

Hamish NC Pike[1][0009-0003-6786-5631]

[1] Department of Molecular Genetics, Weizmann Institute of Science, Rehovot, Israel
`hncpike@gmail.com`

**Abstract.** We investigate the potential of evolving the bytecode of a biologically-inspired virtual machine as a plausible strategy for machine learning. We simulate evolution with the Zyme language and *strand*-based virtual machine. Our test problem is classifying handwritten digits from a subset of the MNIST dataset. Beginning with an initial program with performance no better than random guessing, we achieve consistent accuracy improvements through random mutations over 50 generations. Although these results fall short of state-of-the-art methods like neural networks, they demonstrate that adaptive mutations are found consistently and suggest the potential for evolving Zyme bytecode to competitively tackle the full MNIST task. This result also suggests the value of alternative virtual machines architectures in genetic programming, particularly those optimized for *evolvability*.

**Keywords:** genetic programming, virtual machine, programming language, evolvable bytecode

**Full text:** https://zyme.dev/blog/1_image_classification_by_evolving_bytecode

## 1 Introduction

Despite its conceptual appeal, genetic programming falls far short of modern machine learning methods, particularly neural networks. This disparity is illustrated by the MNIST handwritten digit classification task, which involves identifying the digit (0–9) in 28x28 pixel grayscale images. MNIST is sufficiently straightforward that it serves as an introductory exercise for neural networks, consistently achieving over 99% accuracy, yet remains virtually unsolvable through genetic programming. There have been attempts using hybrid approaches[1-3] however they have yet to demonstrate a complete end-to-end solution: learn directly from raw input data without predefined feature extraction, and evolve complete, self-contained programs that process images and output predictions without relying on external machine learning components.

Traditional approaches rely on program representations derived from C, Java, Python, and other languages used in real-world software development. These languages as well as the computational architectures or virtual machines they run on are optimized for programmability, efficiency, and reliability. Built with a human-centric philosophy, the priority is systems that are predictable and logical, enabling developers to infer program behavior directly from source code while minimizing the risk small mistakes could cause unpredictable and undetectable changes in program behaviour. This rigidity is antithetical to principles of evolution, rendering them fragile to random mutations. Sadly, these human centered design decisions actively undermine the flexibility and adaptability essential for evolutionary processes.

To overcome the limitations of human-oriented languages and competitively solve MNIST, we developed Zyme (https://zyme.dev), an evolution-oriented programming language and virtual machine specifically designed for genetic programming. Zyme is inspired by the molecular processes in living cells. It consists of two components: a language/compiler that generates a bytecode format engineered to be evolvable, and a unique virtual machine built to mimic an abstract cellular metabolism.

Here, we demonstrate the potential of Zyme to solve a small subset of the MNIST dataset, containing only four digits (0, 1, 2, and 3). Beginning with an initial program whose performance is no better than random guessing, we achieve consistent accuracy improvement, accumulating an increase of up to ~40% over 50 generations. Notably, Zyme's instruction set is restricted to only byte-level operations with no built-in image-processing functions5, meaning all progress stems from the system's ability to autonomously derive meaningful features directly from raw binary data. These results suggest that evolution-oriented languages like Zyme could be generalisable across tasks, and if scalable, serve as a foundation for machine learning through genetic programming.

## 2    Results

To illustrate machine learning through the evolution of Zyme programs, we applied the approach to the MNIST dataset for image classification. We will evolve a Zyme program which takes pixel image data (raw binary arrays) as input and returns a predicted label (a single byte) as output. The program's accuracy will be evaluated by testing it on a set of input-output pairs - images with their corresponding labels - and measuring how often it correctly predicts the label.

The current implementation of the Zyme virtual machine is fully functional but remains in a prototype state, making it relatively slow, and limits the scale of preliminary experiments. Thus, we focused on a tiny subset of the MNIST dataset, consisting of four digits (0, 1, 2, and 3), to quickly determine whether there is any potential for learning rather than solving MNIST in its entirety.

Zyme bytecode is evolved as a population of programs, with the starting population consisting of 50 identical copies of a hand-crafted initial program. The evolutionary process follows a two step approach: (1) reproduction with random mutation then (2) selection, where the top performing programs are selected to make up the population of the next generation.

We perform four replicates of this evolutionary process, each running for 50 generations and starting from the same initial population of programs, to assess robustness and reproducibility. Upon completion, we assess the accuracy of each program within every population at each generation using the test dataset (rather than batches of training dataset) to measure their ability to generalize to unseen examples (shown in Figure 1).

Across the replicates, we observe a consistent increase in accuracy, with the final populations of programs achieving accuracy up to ~50%, representing an improvement of 25% compared to the initial random-level performance programs. Although these results do not equate to 'solving' MNIST, they clearly demonstrate that individual mutations in Zyme bytecode can be sufficiently adaptive to yield to measurable performance gain. However, it remains unclear whether these adaptive mutations reflect the exploration of distinct algorithms or are merely trivial adjustments to the initial hand-crafted program, leaving limited potential for further performance improvements.

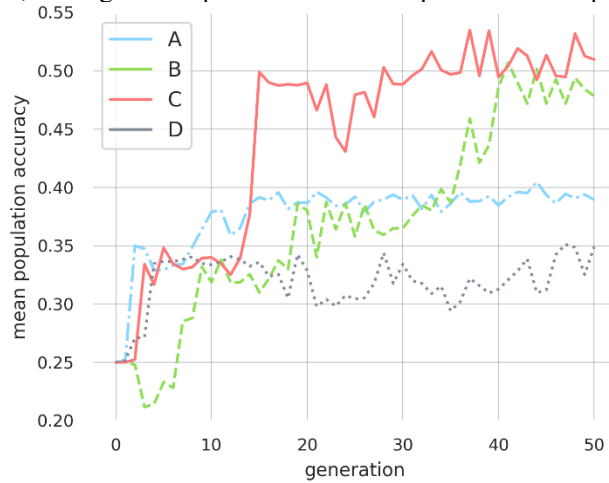

Figure 1 - Line plot showing the mean accuracy of a population of Zyme programs over 50 generations. Each line is a replicate experiment. Accuracy values shown here are evaluated on the unseen test dataset, in contrast to during the selection phase of the artificial evolutionary process where batches of the training dataset are used.

Two of the replicate populations achieve only 35-40% accuracy, highlighting the risk of populations getting stuck and stagnating. Yet, the fact that these populations stabilize at different performance levels suggests that each is uncovering separate mechanisms to enhance performance. Further, the adaptations observed in the more successful populations—those reaching over 50% accuracy—appear to achieve this through incremental gains, where adaptations build on one another. For example, replicate B advance in three steps from 0.25 to 0.33 to 0.37 and up to 0.49. Again the intermediary steps stabilise at unique accuracy levels. This behavior is inconsistent with mutations that merely tweak a few constants in the initial handcrafted program, as such changes would likely cause all populations to converge to similar performance levels.

A closer examination of individual populations (as shown in Figure 2) reveals that some individuals achieve even higher accuracy. For instance, in replicate C, some achieve 65% accuracy, representing a 40% increase over baseline and a 15% improvement compared to the population mean. This indicates that further progress is possible, and the current top population accuracies do not represent the upper limit. However, it remains unclear why the population does not immediately jump to this higher level. We hypothesize that this is due to a mismatch between a program's performance in a specific batch (which determines selection) and its overall performance on the test examples (as shown in the figures). We observe numerous programs (though data is not shown here) that exhibit high test accuracy but relatively low batch accuracy. In such cases, selection does not favor these programs, causing the overall population performance to stall. While this may seem inefficient, we believe it helps prevent overfitting.

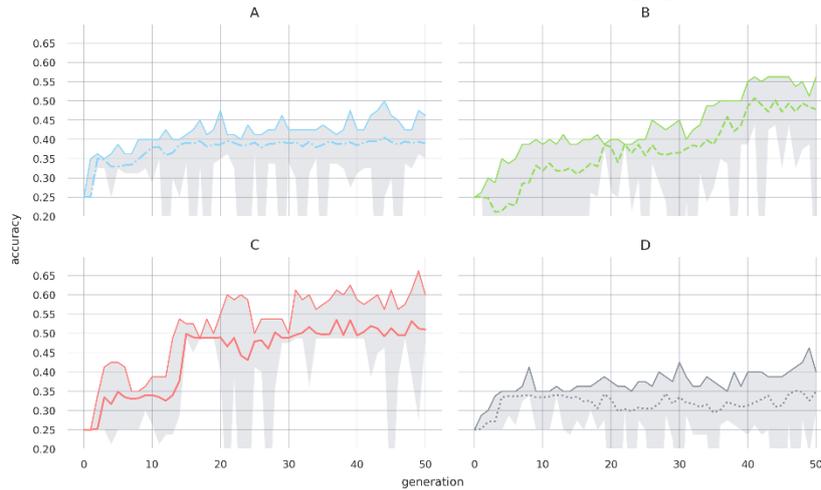

Figure 2 - Line plots showing the distribution of accuracies across populations over 50 generations. Each subplot (a-d) represents an individual replicate. The grey shaded area indicates the range of accuracies, spanning from the minimum to the maximum observed within the population (with the maximum also depicted as the top thin solid line). The central thick lines represent the population mean, consistent with Figure 1.

## 3     Discussion

Genetic programming has long faced a paradoxical question: why can natural evolution produce such remarkable biodiversity while we struggle to evolve computer programs that solve even trivial tasks? Our work suggests the missing element may be specialised computer architectures optimised for evolution. Taking inspiration from biology, we developed Zyme, an *evolution-oriented* language and unique virtual machine.

# Differential Evolution for Optimizing Ensemble Weights in Multiclass Sentiment Classification

Hiram Calvo, Consuelo V. García-Mendoza, Omar J. Gambino, and Miguel G. Villarreal-Cervantes

Instituto Politécnico Nacional, Mexico City, Mexico
**hcalvo@cic.ipn.mx**

**Abstract.** This work addresses the challenge of sentiment polarity classification in an unbalanced multiclass setting. We focus on the Spanish TASS corpus, where the test set is significantly larger than the training set and the class distribution is skewed. Traditional classifiers such as Naive Bayes, Logistic Regression, and SVM perform modestly under these conditions. However, we propose an ensemble approach that uses Differential Evolution to optimize the combination weights, leading to improved performance. Our method outperforms previous baselines on the TASS General corpus without relying on complex deep learning techniques.

**Keywords:** Sentiment analysis · Ensemble learning · Differential Evolution · Unbalanced data · Twitter corpus

## 1 Introduction

Sentiment polarity classification has gained importance due to its applications in analyzing opinions on social media. In particular, datasets like the Spanish TASS corpus pose several challenges: limited training data, a much larger test set, and highly unbalanced class distributions [6, 3, 4, 7]. Traditional machine learning approaches often struggle in this scenario, especially in multiclass settings [6].

While deep learning models such as BERT have achieved excellent results, they demand substantial data and computational resources [7]. We explore a lightweight alternative that combines traditional classifiers through an optimized ensemble strategy. The ensemble is tuned using Differential Evolution (DE), a population-based heuristic known for its robustness in optimizing continuous parameters under constraints [5, 1].

## 2 Evolutionary Optimization of the Weighted Ensemble Classification

Our ensemble integrates three base classifiers: Multinomial Naive Bayes (MNB), Logistic Regression (LR), and Support Vector Machines (SVM). Each classifier outputs a probability distribution over the classes. The ensemble prediction is the weighted average of these distributions:

$$\text{score}(c) = \sum_{i=1}^{n} w_i p_i(c) \tag{1}$$

Where $p_i(c)$ is the probability assigned to class $c$ by classifier $i$, and $w_i$ is the weight assigned to that classifier. The weights are normalized to sum to one.

We use the Differential Evolution (DE) algorithm to search for the optimal weight vector $\boldsymbol{w}$ that maximizes macro-$F_1$ score on a validation set. The DE algorithm evolves a population of weight vectors through mutation, crossover, and selection. It is particularly effective in continuous optimization under noise and complex landscapes.

The DE/rand/1 strategy is defined as follows:

$$\boldsymbol{v} = \boldsymbol{a} + F(\boldsymbol{b} - \boldsymbol{c}) \tag{2}$$

Where $\boldsymbol{a}$, $\boldsymbol{b}$, and $\boldsymbol{c}$ are distinct random vectors from the population, and $F$ is the scaling factor.

The crossover step produces a trial vector $\boldsymbol{u}$ by mixing components of $\boldsymbol{v}$ and the target vector $\boldsymbol{x}$ using crossover probability $CR$. Selection compares the fitness (macro-$F_1$) of $\boldsymbol{u}$ and $\boldsymbol{x}$, and the better one survives.

---

**Algorithm 1** Differential Evolution for Optimizing Ensemble Weights

---

1: Initialize population $P$ with $N$ random weight vectors $\boldsymbol{w}_i \in [0,1]^n$
2: **for** each generation $g = 1$ to $G$ **do**
3:   **for** each individual $\boldsymbol{x}$ in $P$ **do**
4:     Select $\boldsymbol{a}, \boldsymbol{b}, \boldsymbol{c}$ from $P$ randomly and distinct from $\boldsymbol{x}$
5:     Mutate: $\boldsymbol{v} = \boldsymbol{a} + F(\boldsymbol{b} - \boldsymbol{c})$
6:     Crossover: generate trial vector $\boldsymbol{u}$ from $\boldsymbol{v}$ and $\boldsymbol{x}$ using $CR$
7:     Selection: replace $\boldsymbol{x}$ with $\boldsymbol{u}$ if fitness($\boldsymbol{u}$) > fitness($\boldsymbol{x}$)
8:   **end for**
9: **end for**
10: Return best weight vector $\boldsymbol{w}^*$ in $P$

---

This optimization process effectively balances classifier contributions and compensates for class imbalance and skewed distributions.

## 3    Experimental Setup

We used the TASS 2019 General corpus [2], which consists of tweets written in Spanish labeled into four sentiment categories: Positive, Negative, Neutral, and None. The dataset exhibits strong class imbalance. The training set contains 721 tweets, while the test set contains 6,126 tweets. Class distributions differ substantially between training and test sets.

Preprocessing involved lowercasing, removing mentions, hashtags, URLs, and punctuation. We used TF-IDF vectorization with unigrams and bigrams. Stopwords were not removed, as they can carry polarity in short texts [8].

Base classifiers were implemented with scikit-learn using the following configurations: MNB with Laplace smoothing and TF-IDF features; LR with L2 regularization and solver='liblinear'; SVM with linear kernel, $C = 1.0$.

We partitioned 20% of the training data to form a validation set for DE optimization. DE was configured with a population size of 20, 100 generations, crossover probability $CR = 0.9$, and differential weight $F = 0.8$. We used early stopping if no improvement was observed in 20 generations.

In addition to macro-$F_1$, we also evaluated per-class $F_1$ scores and accuracy to ensure balanced performance across categories. The None class, which was particularly underrepresented, benefited most from the weighting optimization.

## 4  Results and Discussion

Table 1 presents the macro-$F_1$ scores obtained on the test set. The DE-optimized ensemble outperformed all individual models as well as the unweighted ensemble.

**Table 1.** Macro-$F_1$ scores on the TASS 2019 test set.

| Method | Macro-$F_1$ |
|---|---|
| Naive Bayes | 0.452 |
| Logistic Regression | 0.468 |
| SVM | 0.475 |
| Unweighted Ensemble | 0.489 |
| DE-Optimized Ensemble | **0.511** |

To analyze convergence behavior, Figure 1 shows the evolution of macro-$F_1$ across generations during the DE optimization process. Performance stabilizes after approximately 60 generations, supporting the use of early stopping. These results demonstrate that DE is well-suited for tuning ensemble weights in constrained and imbalanced scenarios. Unlike deep learning models, our method requires minimal data and no pretraining.

Furthermore, the ensemble approach shows more stable performance under variations in class distributions, making it more robust in dynamic social media environments. Additional experiments with alternative base classifiers, such as decision trees and random forests, are under exploration.

## 5  Conclusion

We presented an efficient ensemble method for sentiment classification on unbalanced multiclass corpora. Our approach combines traditional classifiers through Differential Evolution, which optimizes their weights to improve macro-$F_1$ score.

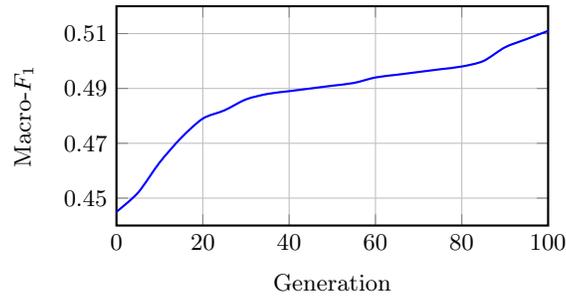

**Fig. 1.** Convergence curve of macro-$F_1$ during DE optimization.

Without relying on deep learning or feature-rich representations, we achieved competitive performance on the TASS benchmark. Future work includes integrating dynamic ensemble selection, testing with different optimization strategies (e.g., PSO, CMA-ES), and extending the method to multilingual datasets and low-resource settings.

# Using SHAP to visualize Pre-Match Outcome Prediction in Dota 2

Julio Losada-Rodríguez[1][0009−0008−5079−0422], Pedro A.
Castillo[1][0000−0002−5258−0620], Antonio Mora[2][0000−0003−1603−9105], and Pablo
García-Sánchez[1][0000−0003−4644−2894]

[1] Department of Computer Engineering, Automation and Robotics, CITIC-UGR,
University of Granada, 18014 Granada, Spain
julio.losada.rodriguez@gmail.com,pablogarcia@ugr.es,pacv@ugr.es
[2] Department of Signal Theory, Telematics and Communication CITIC-UGR,
University of Granada, 18014 Granada, Spain
amorag@ugr.es

**Abstract.** Predicting outcomes in competitive video games like Dota 2 is crucial for strategic planning in e-sports. This study employs machine learning (ML) models—Random Forest, Gradient Boosting, and Logistic Regression—to predict match outcomes based solely on hero selection data. Using the OpenDota API, we collected and preprocessed 4,500 matches, evaluating models via accuracy, precision, recall, and F1-score. Random Forest achieved 98% accuracy, outperforming other models. Explainability techniques, particularly SHAP (SHapley Additive exPlanations), revealed key heroes influencing predictions. This work highlights the potential of explainable AI (XAI) in gaming, offering insights for players and developers to optimize strategies.

**Keywords:** videogames · machine learning · explainability · soft computing · MOBA.

## 1 Introduction

Over the last decade, Multiplayer Online Battle Arena (MOBA) video games have emerged as a dominant force in the digital entertainment industry, capturing the attention of millions of gamers around the world. Among these, *Dota 2*, developed by Valve Corporation, has established itself as one of the most iconic and competitive titles of the genre. MOBA games, characterized by matches in which teams of players control heroes with unique abilities, facing each other in arenas with the goal of destroying the enemy's base, have revolutionized the way video games are conceived and enjoyed.

The goal of this study is not only to discuss how machine learning models can be used to predict game outcomes in this scope, but also to introduce explainable AI (XAI) techniques [1] as an innovative element. While many predictive models in the field of video games focus exclusively on improving the accuracy of predictions [2,3], this paper puts special emphasis on the interpretability of the models, allowing us to understand what factors or variables have a higher influence on the predictions. Incorporating explainability not only helps to identify the reasons behind a win or loss, but also allows players, developers, and game designers to gain valuable insight into the behaviour of the model, facilitating a deeper understanding of game dynamics and fostering confidence in the predictions generated.

## 2   Experiments and results

We extracted 4,500 matches from OpenDota API [5]. Each record included 10 hero IDs (5 per team) and a binary outcome (Radiant/Dire victory). One-hot codification was used.

The next models were compared: **Random Forest** (RF): 100 estimators, no max depth. **Gradient Boosting** (GB): 100 estimators, learning rate = 0.1. **Logistic Regression** (LR): Max iterations = 1,000. The metrics used were Accuracy, precision, recall, F1-score, AUC-ROC, with a train/test split of 80%/20% with cross-validation. Results are shown in Table 1.

Table 1: Summary of Results for the ML Algorithms

| Metric | Random Forest | Gradient Boosting | Logistic Regression |
|---|---|---|---|
| Accuracy (Train/Test Split) | 0.9800 | 0.8955 | 0.6588 |
| Accuracy (Cross-Validation) | 0.9837 | 0.8653 | 0.6562 |
| Precision | 0.9795 | 0.8916 | 0.6656 |
| Recall | 0.9869 | 0.9383 | 0.8560 |
| F1-Score | 0.9832 | 0.9143 | 0.7489 |

It is important to note that the focus of this article is not mainly on the results of these methods, but it is on the analysis of explainability. SHAP (SHapley Additive exPlanations) leverages game-theoretic Shapley values to quantify feature contributions in ML models, offering both local and global interpretability [4]. Each feature's SHAP value reflects its impact: positive values increase prediction likelihood, while negative values decrease it.

In order to interpret the graphics:

- Feature Importance – A wider SHAP value range indicates greater predictive influence.
- Contribution Direction – Chart position and color (e.g., red/blue) denote whether features enhance or reduce the outcome probability.
- Effect Distribution – Horizontal dispersion reflects variability in a feature's impact across predictions.

## 3   Conclusions

In this paper, we have explored the possibility of predicting the outcome of Dota 2 games based solely on character selection prior to the start of the game. To do so, we used three machine learning algorithms, namely Logistic Regression, Random Forest and Gradient Boosting, and trained them on historical data obtained through the OpenDota API from real matches on the game.

However, despite the positive results, the study has some limitations: the model is based only on hero selection, without considering other dynamic variables that occur during the game, such as real-time strategic decisions, individual player skill, and in-game events. These variables will be addressed in future work. In any case, SHAP charts have been used to analyse the importance of the features, showing the 5 most influential features in each case, in order to see which were the 5 heroes that most influenced the model in predicting a win for the team.

The results of this study lead us to think that similar methodologies could be applied in the future to other video games and e-sports, integrating additional contextual variables to improve prediction accuracy and optimize pregame strategic decision making.

**Acknowledgments.** This research has been funded by the Ministerio Español de Ciencia e Innovación under project number PID2023-147409NB-C21 funded by MICIU/AEI/10.13039/501100011033.

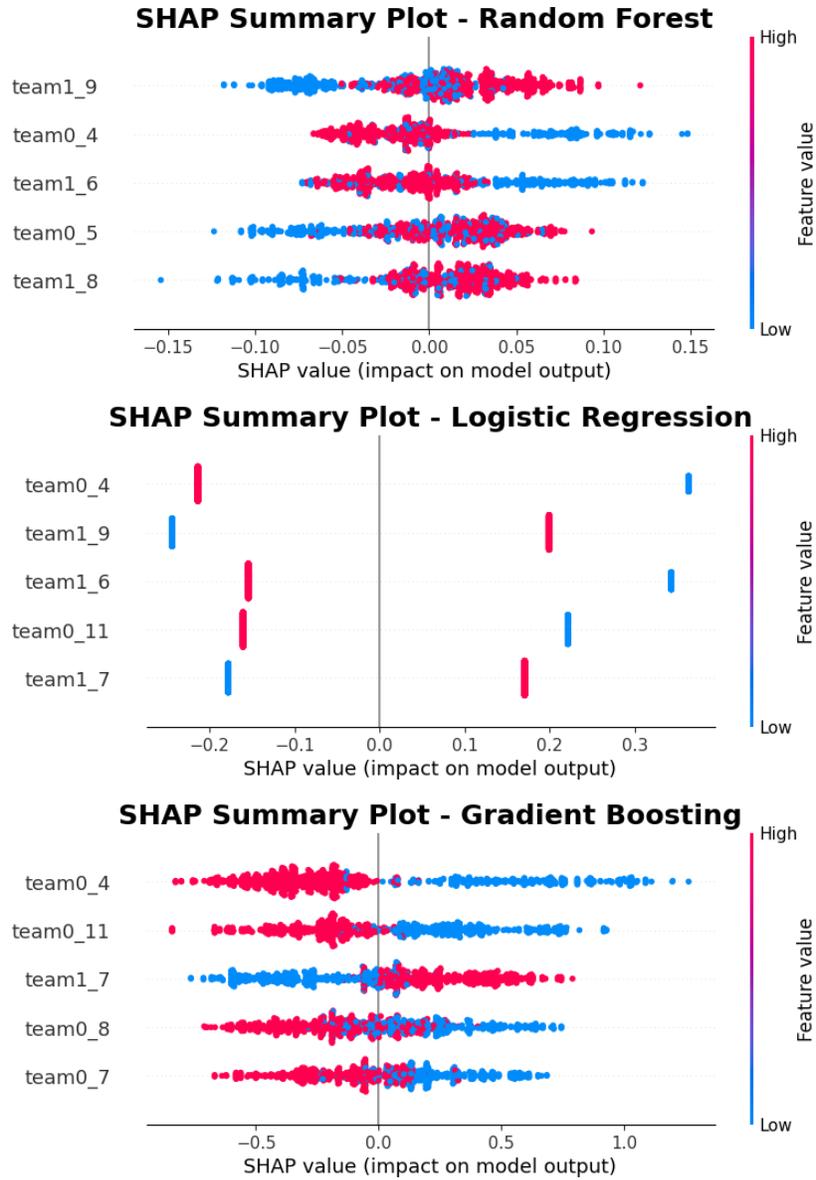

Fig. 1: SHAP values of the 5 heroes with the greatest impact on the different models